\documentclass[journal,draftclsnofoot]{IEEEtran}
\onecolumn
\usepackage{latexsym}
\usepackage{cite}
\usepackage{amssymb}
\usepackage{bbm}
\usepackage{textpos}
\usepackage{enumerate}
\usepackage{bm}
\usepackage{mathtools}
\usepackage[usenames,dvipsnames]{xcolor}
\usepackage{stmaryrd}
\usepackage{amsmath, amssymb,accents}
\usepackage{dsfont}
\usepackage{caption}
\usepackage{algorithmicx}
\usepackage{algorithm} 
\usepackage[noend]{algpseudocode} %
\usepackage{comment}
\usepackage{accents}

\DeclareMathOperator{\relu}{ReLU}
\DeclareMathOperator{\sig}{sigmoid}
\usepackage{graphicx}
\usepackage{psfrag}
\usepackage{units}
\usepackage{subfig}
\usepackage{setspace}
\usepackage{theorem}
\usepackage{mathtools}
\usepackage{fancyhdr}
\allowdisplaybreaks

\newtheorem{definition}{Definition}

\newtheorem{theorem}{Theorem}

\newtheorem{proposition}{Proposition}

\newcommand\numberthis{\addtocounter{equation}{1}\tag{\theequation}}
{} 

\begin{document}

\makeatletter
\newcommand{\figCaption}[2]{\caption[#1]{#1}{#2}}
\DeclareRobustCommand{\rchi}{{\mathpalette\irchi\relax}}
\newcommand{\irchi}[2]{\raisebox{\depth}{$#1\chi$}} 
\newcommand{\cA}{\mathcal{A}}
\newcommand{\cB}{\mathcal{B}}
\newcommand{\cC}{\mathcal{C}}
\newcommand{\cD}{\mathcal{D}}
\newcommand{\cE}{\mathcal{E}}
\newcommand{\cF}{\mathcal{F}}
\newcommand{\cG}{\mathcal{G}}
\newcommand{\cH}{\mathcal{H}}
\newcommand{\cI}{\mathcal{I}}
\newcommand{\cJ}{\mathcal{J}}
\newcommand{\cK}{\mathcal{K}}
\newcommand{\cL}{\mathcal{L}}
\newcommand{\cM}{\mathcal{M}}
\newcommand{\cN}{\mathcal{N}}
\newcommand{\cO}{\mathcal{O}}
\newcommand{\cP}{\mathcal{P}}
\newcommand{\cQ}{\mathcal{Q}}
\newcommand{\cR}{\mathcal{R}}
\newcommand{\cS}{\mathcal{S}}
\newcommand{\cT}{\mathcal{T}}
\newcommand{\cU}{\mathcal{U}}
\newcommand{\cV}{\mathcal{V}}
\newcommand{\cW}{\mathcal{W}}
\newcommand{\cX}{\mathcal{X}}
\newcommand{\cY}{\mathcal{Y}}
\newcommand{\cZ}{\mathcal{Z}}
\newcommand{\FSG}{\mathcal{F}^{(\mathsf{SG})}_{d,K}}
\newcommand{\BB}{\mathbb{B}}
\newcommand{\CC}{\mathbb{C}}
\newcommand{\DD}{\mathbb{D}}
\newcommand{\EE}{\mathbb{E}}
\newcommand{\FF}{\mathbb{F}}
\newcommand{\GG}{\mathbb{G}}
\newcommand{\HH}{\mathbb{H}}
\newcommand{\II}{\mathbb{I}}
\newcommand{\JJ}{\mathbb{J}}
\newcommand{\KK}{\mathbb{K}}
\newcommand{\LL}{\mathbb{L}}
\newcommand{\MM}{\mathbb{M}}
\newcommand{\NN}{\mathbb{N}}
\newcommand{\OO}{\mathbb{O}}
\newcommand{\PP}{\mathbb{P}}
\newcommand{\QQ}{\mathbb{Q}}
\newcommand{\RR}{\mathbb{R}}
\newcommand{\TT}{\mathbb{T}}
\newcommand{\UU}{\mathbb{U}}
\newcommand{\VV}{\mathbb{V}}
\newcommand{\WW}{\mathbb{W}}
\newcommand{\XX}{\mathbb{X}}
\newcommand{\YY}{\mathbb{Y}}
\newcommand{\ZZ}{\mathbb{Z}}
\newcommand*{\dd}{\, \mathrm{d}}
\newcommand{\vasti}{\bBigg@{3.5 }}
\newcommand{\vast}{\bBigg@{4}}
\newcommand{\Vast}{\bBigg@{5}}
\newcommand{\Vastt}{\bBigg@{7}}
\makeatother
\newcommand{\be}{\begin{equation}}
\newcommand{\ee}{\end{equation}}
\newcommand{\ba}{\begin{align}}
\newcommand{\ea}{\end{align}}
\newcommand{\baa}{\begin{align*}}
\newcommand{\eaa}{\end{align*}}
\newcommand{\ber}{$\ \mbox{Ber}$}
\newcommand{\argmin}{\mathop{\mathrm{argmin}}}
\newcommand{\argmax}{\mathop{\mathrm{argmax}}}
\newcommand{\ubar}[1]{\underaccent{\bar}{#1}}
\newcommand*{\gauss}{\varphi_\sigma}
\newcommand*{\gausss}{\varphi_{\frac{\sigma}{\sqrt{2}}}}
\newcommand*{\gaussI}{\varphi_1}
\newcommand*{\gausssI}{\varphi_{\frac{1}{\sqrt{2}}}}
\newcommand*{\Gauss}{\mathcal{N}_\sigma}
\newcommand*{\Gausss}{\mathcal{N}_{\frac{\sigma}{\sqrt{2}}}}
\newcommand*{\dkl}{\mathsf{D}_{\mathsf{KL}}}
\newcommand{\p}[1]{\left(#1\right)}
\newcommand{\s}[1]{\left[#1\right]}
\newcommand{\cp}[1]{\left\{#1\right\}}
\newcommand{\abs}[1]{\left|#1\right|}
\newcommand{\nmc}{n_{\mathsf{MC}}}
\newcommand{\var}{\mathsf{var}}
\newcommand*{\E}{\mathbb E}
\newcommand*{\Chi}{\rchi^2}
\newcommand{\fillater}[1] {{\Large \color{red}[#1]}}

\def\yp#1{{\color{red}YP: #1}}
\def\eqdef{\triangleq}

\makeatletter
\def\Ddots{\mathinner{\mkern1mu\raise\p@
\vbox{\kern7\p@\hbox{.}}\mkern2mu
\raise4\p@\hbox{.}\mkern2mu\raise7\p@\hbox{.}\mkern1mu}}
\makeatother

\title{The Information Bottleneck Problem and Its Applications in Machine Learning}

\author{Ziv Goldfeld and Yury Polyanskiy

		\thanks{
		Z. Goldfeld is with the Electrical and Computer Engineering Department, Cornell University, Ithaca, NY, 14850, US (e-mail: goldfeld@cornell.edu). Y. Polyanskiy is with the Department of Electrical Engineering and Computer Science, Massachusetts Institute of Technology, Cambridge, MA, 02139, US (e-mail: yp@mit.edu).}
}
\maketitle


\begin{abstract}





Inference capabilities of machine learning (ML) systems skyrocketed in recent years, now playing a pivotal role in various aspect of society. The goal in statistical learning is to use data to obtain simple algorithms for predicting a random variable $Y$ from a correlated observation $X$. Since the dimension of $X$ is typically huge, computationally feasible solutions should summarize it into a lower-dimensional feature vector $T$, from which $Y$ is predicted. The algorithm will successfully make the prediction if $T$ is a good proxy of $Y$, despite the said dimensionality-reduction. A myriad of ML algorithms (mostly employing deep learning (DL)) for finding such representations $T$ based on real-world data are now available. While these methods are often effective in practice, their success is hindered by the lack of a comprehensive theory to explain it. The information bottleneck (IB) theory recently emerged as a bold information-theoretic paradigm for analyzing DL systems. Adopting mutual information as the figure of merit, it suggests that the best representation $T$ should be maximally informative about $Y$ while minimizing the mutual information with $X$. In this tutorial we survey the information-theoretic origins of this abstract principle, and its recent impact on DL. For the latter, we cover implications of the IB problem on DL theory, as well as practical algorithms inspired by it. Our goal is to provide a unified and cohesive description. A clear view of current knowledge is particularly important for further leveraging IB and other information-theoretic ideas to study
DL models.




\end{abstract}


\section{Introduction}



The past decade cemented the status of machine learning (ML) as a method of choice for a variety of inference tasks. The general learning problem considers an unknown probability distribution $P_{X,Y}$ that generates a target variable $Y$ and an observable correlated variable $X$. Given a dataset from $P_{X,Y}$, the goal is to learn a representation $T(X)$ of $X$ that is useful for inferring $Y$.\footnote{The reader is referred to~\cite{hastie2009elements} for background on statistical learning theory.} While these ideas date back to the late 1960's~\cite{vapnik1968uniform,minsky1969perceptrons}, it was not until the 21st century that ML, and specifically, deep learning (DL), began to revolutionize data science practice~\cite{lecun2015deep}. Fueled by increasing computational power and data availability, deep neural networks (DNNs) are often unmatched in their effectiveness for classification, feature extraction and generative modeling. 

This practical success, however, is not coupled with a comprehensive theory to explain how and why deep models work so well on real-world data. In recent years, information theory emerged as a popular lens through which to study fundamental properties of DNNs and their learning dynamics~\cite{tishby_DNN2015,DNNs_Tishby2017,Achille2018,achille2018information,DNNs_ICLR2018,gabrie2018entropy,yu2018understanding,cheng2018evaluating,ICML_Info_flow2019,wickstrom2019information,amjad2019learning}. In particular, the information bottleneck (IB) theory~\cite{tishby_DNN2015,DNNs_Tishby2017} received significant attention. In essence, the theory extends an information-theoretic framework introduced in 1999~\cite{tishby_IB1999} to account for modern DL systems~\cite{tishby_DNN2015,DNNs_Tishby2017}. It offers a novel paradigm for analyzing DNNs in an attempt to shed light on their layered structure, generalization capabilities and learning dynamics.




The IB theory for DL includes claims about the computational benefit of deep architectures, that generalization is driven by the amount of information compression the network attains, and more\cite{tishby_DNN2015,DNNs_Tishby2017}. This perspective inspired many followup works~\cite{DNNs_ICLR2018,amjad2019learning,ICML_Info_flow2019,yu2018understanding,cheng2018evaluating,wickstrom2019information,goldfeld2019convergence}, some corroborating (and building on) and others refuting different aspects of the theory. This tutorial surveys the information-theoretic origins of the IB problem and its recent applications to and influences on DL. Our goal is to provide a comprehensive description that accounts for the multiple facets of this opinion-splitting topic.

\subsection{Origins in Information Theory}


The IB problem was introduced in~\cite{tishby_IB1999} as an information-theoretic framework for learning. It considers extracting information about a target signal $Y$ through a correlated observable $X$. The extracted information is quantified by a variable $T=T(X)$, which is (a possibly randomized) function of $X$, thus forming the Markov chain $Y\leftrightarrow X\leftrightarrow T$. The objective is to find a $T$ that minimizes the mutual information $I(X;T)$, while keeping $I(Y;T)$ above a certain threshold; the threshold determines how informative the representation $T$ is of $Y$. Specifically, the IB problem for a pair $(X,Y)$ (with a know joint probability law) is given by 
\begin{align*}
\inf\ \ &I(X;T)\\
\mbox{subject to: }\  &I(Y;T)\geq \alpha,  
\end{align*}
where minimization is over all randomized mappings of $X$ to $T$. This formulation provides a natural \emph{approximate} version of minimal sufficient statistic (MSS)~\cite{lehmann1955completeness} (cf.~\cite{kullback1951information}). Notably, the same problem was introduced back in 1975 by Witsenhausen and Wyner~\cite{WW75} in a different context: as a tool for analyzing common information~\cite{GK73}.



Beyond formulating the problem,~\cite{tishby_IB1999} showed that, when alphabets are discrete, optimal $T$ can be found by iteratively solving a set of self-consistent equations. An algorithm (a generalization of Blahut-Arimoto~\cite{Blahut72,Arimoto72}) for solving the equations was also provided. Extensions of the IB problem and the aforementioned algorithm to the distributed setup were given in \cite{aguerri2018distributeddiscrete}. The only continuous-alphabet adaptation of this algorithm is known for jointly Gaussian $(X,Y)$~\cite{tishby_Gaussian_IB2005} (cf. \cite{estella2018distributedgaussian} for the distributed Gaussian IB problem). In this case, the solution reduces to a canonical correlation analysis (CCA) projection with tunable rank. Solving the IB problem for complicated $P_{X,Y}$ distributions seems impossible (even numerically), even more so when only samples of $P_{X,Y}$ are available. The IB formulation, its solution, relations to MSSs, and the Gaussian case study are surveyed in Section~\ref{SEC:IB}. 

Optimized information measures under constraints often appear in information theory as solutions to operational coding problems. It is thus natural to ask whether there is an operational setup whose solution is the IB problem. Indeed, the classic framework of remote source coding (RSC) fits this description. RSC considers compressing a source $X^n$ so as to recover a correlated sequence $Y^n$ from that compression, subject to a distortion constraint~\cite{dobrushin1962information,wolf1970transmission}. Choosing the distortion measure as the logarithmic loss and properly setting the compression threshold, recovers the IB problem as the fundamental RSC rate. With that choice of loss function, RSC can roughly be seen as a multi-letter extension of the IB problem. The connection between the two problems is covered in Section~\ref{SEC:remote_sc}. The reader is referred to \cite{zaidi2020tutorial} for a recent survey covering additional connections between the IB problem and other information-theoretic setups.


\subsection{The Information Bottleneck Theory for Deep Learning}

Although first applications of the IB problem to ML, e.g., for clustering~\cite{slonim2000agglomerative}, date two decades ago, it recently gained much traction in the context of DL. From the practical perspective, the IB principle was adopted as a design tool for DNN classifiers~\cite{alemi2017deep,achille2018information} and generative models~\cite{higgins2017beta}, with all three works published concurrently. Specifically,~\cite{alemi2017deep} optimized a DNN to solve the IB Lagrangian via gradient-based methods. Termed deep \emph{variational IB} (VIB), the systems learns stochastic representation rules that were shown to generalize better and enjoy robustness to adversarial examples. The same objective was studied in~\cite{achille2018information}, who argued it promotes minimality, sufficiency and disentanglement of representations. The disentanglement property was also employed in~\cite{higgins2017beta} for generative modeling purposes, where the $\beta$-variational autoencoder was developed. These applications of the IB framework are covered in Section~\ref{SUBSEC:IB_practice}.
\ \\

The IB problem also had impact on DL theory. The main characterizing property of DNNs, as compared to general learning algorithms, is their layered structure. The IB theory suggests that deeper layers correspond to smaller $I(X;T)$ values, thus providing increasingly compressed sufficient statistics. In a classification task, the feature $X$ might contain information that is redundant for determining the label $Y$. It is therefore desirable to find representations $T$ of $X$ that shed redundancies, while retaining informativeness about $Y$. The argument of~\cite{tishby_DNN2015,DNNs_Tishby2017} was that the IB problem precisely quantifies the fundamental tradeoff between informativeness (of $T$ for $Y$) and compression (of $X$ into $T$). 

the IB theory for DL was first presented in~\cite{tishby_DNN2015}, followed by the supporting empirical study~\cite{DNNs_Tishby2017}. The latter relied on a certain synthetic binary classification task as a running example. Beyond testing claims made in~\cite{tishby_DNN2015}, the authors of~\cite{DNNs_Tishby2017} further evolved the IB theory. A main new argument was that classifiers trained with cross-entropy loss and stochastic gradient descent (SGD) inherently (try to) solve the IB optimization problem. As such,~\cite{DNNs_Tishby2017} posed the \emph{information plane} (IP), i.e., the trajectory in $\RR^2$ of the mutual information pair $\big(I(X;T),I(Y;T)\big)$ across training epochs, as a potent lens through which to analyze DNNs.


Based on this IP analysis, the IB theory proposed certain predictions about DNN learning dynamics. Namely,~\cite{DNNs_Tishby2017} argued that there is an inherent phase transition during DNN training, starting from a quick `fitting' phase that aims to increase $I(Y;T)$ and followed by a long `compression' phase that shrinks $I(X;T)$. This observation was explained as the network shedding redundant information and learning compressed representations, as described above. This is striking since the DNN has no explicit mechanism that encourages compressed representations. The authors of~\cite{DNNs_Tishby2017} further used the IP perspective to reason about computational benefits of deep architectures, phase transitions in SGD dynamics, and the network's generalization error. The IB theory for DL is delineated in section~\ref{SUBSEC:IB_theory}. 


\subsection{Recent Results and Advances}

The bold IB perspective on DL~\cite{tishby_DNN2015,DNNs_Tishby2017} inspired followup research aiming to test and understand the observations therein. In~\cite{DNNs_ICLR2018}, the empirical findings from~\cite{DNNs_Tishby2017} were revisited. The goal was to examine their validity in additional settings, e.g., across different nonlinearities, replacing SGD with batch gradient-descent, etc. The main conclusion of~\cite{DNNs_ICLR2018} was that the findings from~\cite{DNNs_Tishby2017} do not hold in general. Specifically,~\cite{DNNs_ICLR2018} provided experiments showing that IP dynamics undergo a compression phase only when double-sided saturating nonlinearities (e.g., $\tanh$, as used in~\cite{DNNs_Tishby2017}) are employed. Retraining the same network but with $\relu$ activations, produces a similarly-performing classifier whose IP trajectory does \emph{not exhibit compression}. This refuted the fundamental role of compression in learning deep models, as posed in~\cite{DNNs_Tishby2017}. The results of \cite{DNNs_ICLR2018}, along with additional  counterexamples they provided to claims from~\cite{DNNs_Tishby2017}, are covered in Section~\ref{SUBSEC:ICLR}.

Theoretical aspect of the IB theory for DL were also reexamined. It was noted in
\cite{ICML_Info_flow2019,amjad2019learning} that the mutual information measures of interest $\big(I(X;T),I(Y;T)\big)$ are vacuous in deterministic DNNs (i.e., networks that define a deterministic mapping for each fixed set of parameters). This happens because deterministic DNNs with strictly monotone activations (e.g., $\tanh$ or $\sig$) can encode the entire input dataset into arbitrarily fine variations of $T$. Consequently, no information about $X$ is lost when traversing the network's layers and $I(X;T)$ is either the \emph{constant} $H(X)$ (discrete $X$) or \emph{infinite} (continuous $X$); a vacuous quantity, independent of the network parameters, either way~\cite{ICML_Info_flow2019}.\footnote{A similar degeneracy occurs for $I(Y;T)$, which, e.g., equals $I(Y;X)$, whenever $X$ is discrete and the activations are injective. Note that, except for synthetic models, it is customary to replace the unknown joint distribution $P_{X,Y}$ with its empirical version, thus making $X$ discrete.} Similar degeneracies occur in DNNs with any bi-Lipschitz nonlinearities, as well as for $\relu$s~\cite{amjad2019learning}. This implies that the (estimated) IP trajectories presented in~\cite{DNNs_Tishby2017,DNNs_ICLR2018} for deterministic DNNs, cannot be fluctuating (e.g., undergoing fitting/compression phases) due to changes in mutual information. Indeed, \cite{ICML_Info_flow2019} showed that these fluctuations are an artifact of the quantization-based method employed in \cite{DNNs_Tishby2017,DNNs_ICLR2018} to approximate the true mutual information. We describe the technical caveats in applying information-theoretic reasoning to deterministic DL systems and the misestimation of mutual information in Sections~\ref{SUBSUBSEC:IB_vacuous} and  \ref{SUBSEC:misestimation}, respectively.

To circumvent these issues,~\cite{ICML_Info_flow2019} proposed an auxiliary framework, termed \emph{noisy DNNs}, over which $I(X;T)$ and $I(Y;T)$ are meaningful, parameter-dependent quantities. In this framework, additive Gaussian noise is injected to the output of each of the network's neurons.~\cite{ICML_Info_flow2019} showed that noisy DNNs approximate deterministic ones both in how they generalize (just as well and sometimes better), and in terms of the learned representations. From a theoretical standpoint, the noise renders $X\mapsto T$ a stochastic parametrized channel, which makes $I(X;T)$ and $I(Y;T)$ functions of the network's parameters. Once the degeneracy of the considered mutual information terms was alleviated,~\cite{ICML_Info_flow2019} focused on accurately measuring their values across training epochs.

Building on recent results on differential entropy estimation under Gaussian smoothing~\cite{Goldfeld_estimation_ICSEE2019,Goldfeld_estimation_ISIT2019,goldfeld2019convergence}, a rate-optimal estimator of $I(X;T)$ over noisy DNNs was developed~\cite{ICML_Info_flow2019}. This enabled tracking $I(X;T)$ across training of noisy DNN classifiers and empirically show that it also undergoes compression (just like the quantization-based estimate of mutual information over deterministic networks). To understand the relation between compression and the geometry of latent representations,~\cite{ICML_Info_flow2019} described an analogy between $I(X;T)$ and data transmission over additive white Gaussian noise (AWGN) channels. The analogy gave rise to an argument that compression of $I(X;T)$ is driven by the progressive clustering of internal representations of equilabeled inputs.

Armed with the connection between compression and clustering,~\cite{ICML_Info_flow2019} traced back to deterministic DNNs and showed that, while compression of $I(X;T)$ is impossible therein, these networks nonetheless cluster equilabeled samples. They further demonstrated that the quantization-based proxy of mutual information used in~\cite{DNNs_Tishby2017,DNNs_ICLR2018} in fact measures clustering in the latent spaces. This identified clustering of representations as the fundamental phenomenon of interest, while elucidating some of the machinery DNNs employ for learning. Section~\ref{SUBSEC:IB_clustering} elaborates on noisy DNNs, mutual information estimation, and the relation to clustering.  

The IB problem remains an active area of research in ML and beyond. Its rich and opinion-splitting history inspired a myriad of followup works aiming to further explore and understand it. This tutorial surveys a non-exhaustive subset of these works as described above\footnote{Others relevant papers include~\cite{strouse2017deterministic,higgins2017beta,Achille2018,yu2018understanding,cheng2018evaluating,wickstrom2019information,cvitkovic2019minimal}.}, that contributed to different facets of IB research, both historically and more recently. We aim to provide a balanced description that clarifies the current status of the IB problem applied to DL theory and practice. Doing so would help further leveraging the IB and other information-theoretic concepts for the study of DL systems.

\section{Information Bottleneck Formulation and Gaussian Solution}\label{SEC:IB}

The IB framework was proposed in~\cite{tishby_IB1999} as a principled approach for extracting `relevant' or `useful' information from an observed signal about a target one. Consider a pair of correlated random variables $(X,Y)$, where $X$ is the observable and $Y$ is the object on interest. The goal is to compress $X$ into a representation $T$ that preserves as much information about $Y$ as possible.

The formulation of this idea uses mutual information. For a random variable pair $(A,B)\sim P_{A,B}$ with values in $\cA\times\cB$, set
\[
I(A;B):=\int_{\cA\times\cB}\log\left(\frac{\dd P_{A,B}}{\dd P_A\otimes P_B }(a,b)\right)\dd P_{A,B}(a,b),
\]
as the mutual information between $A$ and $B$, where $\frac{\dd P_{A,B}}{\dd P_A\otimes P_B }$ is the Radon-Nikodym derivative of $P_{A,B}$ with respect to (w.r.t.) $P_A\otimes P_B$. Mutual information is a fundamental measure of dependence between random variables with many desirable properties. For instance, it nullifies if and only if $A$ and $B$ are independent, and is invariant to bijective transformations. In fact, mutual information can be obtained axiomatically as a unique (up to a multiplicative constant) functional satisfying several natural `informativeness' conditions~\cite{CovThom06}.


\subsection{The Information Bottleneck Framework} 

The IB framework for extracting the relevant information an $\cX$-valued random variable $X$ contains about a $\cY$-valued $Y$ is described next. For a set $\cT$, let $P_{T|X}$ be a transition kernel from $\cX$ to $\cT$.\footnote{Given two measurable spaces $(\cA,\mathfrak{A})$ and $(\cB,\mathfrak{B})$, $P_{A|B}(\cdot|\cdot):\mathfrak{A}\times\cB\to \RR$ is a transition kernel from the former to the latter if $P_{A|B}(\cdot|b)$ is a probability measure on $(\cA,\mathfrak{A})$, for all $b\in\cB$, and $P_{A|B}(a|\cdot)$ is $\mathfrak{B}$-measurable for all $a\in\mathfrak{A}$.} The kernel $P_{T|X}$ can be viewed as transforming $X\sim P_X$ (e.g., via quantization) into a representation of $T\sim P_T(\cdot):=\int P_{T|X}(\cdot|x)\dd P_X(x)$ in the $\cT$ space. The triple $Y\leftrightarrow X\leftrightarrow T$ forms a Markov chain in that order w.r.t. the joint probability measure $P_{X,Y,T}=P_{X,Y}P_{T|X}$. This joint measure specifies the mutual information terms $I(X;T)$ and $I(Y;T)$, that are interpreted as the amount of information $T$ conveys about $X$ and $Y$, respectively. 

\begin{figure}[!t]
	\begin{center}
	        \includegraphics[scale = .6]{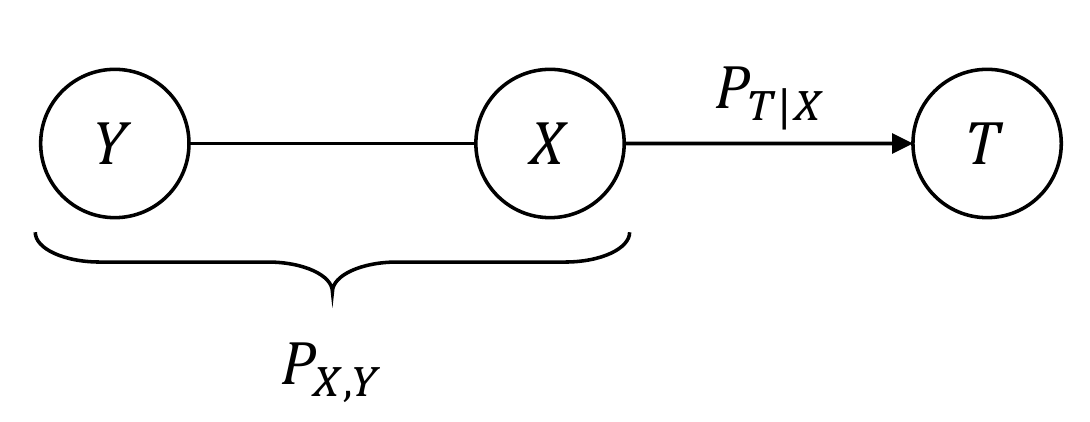}
	        \caption{Graphical representation of probabilistic relations in the IB framework. The triple $(X,Y,T)$ is jointly distributed according to $P_{X,Y}P_{T|X}$, thus forming a Markov chain $Y\leftrightarrow X\leftrightarrow T$. The goal is to find a compressed representation $T$ of $X$ (via the transition kernel $P_{T|X}$), i.e., minimize $I(X;T)$, while preserving at least $\alpha$ bits of information about $Y$, i.e., $I(Y;T)\geq \alpha$.} \label{FIG:IB_block_diagram}
	\end{center}
\end{figure}

The IB framework concerns finding a $P_{T|X}$ that extracts information about $Y$, i.e., high $I(Y;T)$, while maximally compressing $X$, which is quantified as keeping $I(X;T)$ small. Since the Data Processing Inequality (DPI) implies $I(Y;T)\leq I(X;Y)$, the compressed representation $T$ cannot convey more information than the original signal. This gives rise to a tradeoff between compressed representations and preservation of meaningful information. The tradeoff is captured by a parameter $\alpha\in\RR_{\geq 0}$ that specifies the lowest tolerable $I(Y;T)$ values. Accordingly, the IB problem is formulated through the constrained optimization
\begin{equation}
    \inf_{P_{T|X}:\ I(Y;T)\geq \alpha} I(X;T),\label{EQ:IB_contrained_opt}
\end{equation}
where the underlying joint distribution is $P_{X,Y}P_{T|X}$. Thus, we pass the information that $X$ contains about $Y$ through a `bottleneck' via the representation $T$ (see Fig.~\ref{FIG:IB_block_diagram}). An extension of the IB problem to the distributed setup was proposed in \cite{estella2018distributedgaussian,aguerri2018distributeddiscrete}. In that problem, multiple sources $X_1,\ldots,X_K$ are compressed separately into representations $T_1,\ldots,T_K$, that, collectively, preserve as much information as possible about $Y$.

A slightly different form of the problem in \eqref{EQ:IB_contrained_opt}, namely
\begin{equation}
    \inf_{P_{T|X}:\ H(X|T)\geq \alpha} H(Y|T),\label{EQ:WW75}
\end{equation}
first appeared in a seminal paper of Witsenhausen and Wyner~\cite{WW75}, who used it to simplify the proof of G\'acs and K\"orner's result on common information~\cite{GK73}.

\subsection{Lagrange Dual Form and Information Bottleneck Curve}

Note that the optimization problem~\eqref{EQ:IB_contrained_opt} is not convex (in $P_{T|X}$), however it can be made into a convex rate-distortion problem (cf.~\eqref{eq:rd_rsc2}-\eqref{EQ:RSC_RD}). In any case, \eqref{EQ:IB_contrained_opt} is commonly solved by introducing the Lagrange multiplier $\beta$ and considering the functional
\begin{equation}
    \cL_\beta(P_{T|X}):= I(X;T)-\beta I(T;Y).\label{EQ:IB_lagrange}
\end{equation}
With this definition, the IB problem can be recast as minimizing $\cL_\beta(P_{T|X})$ over all possible $P_{T|X}$ kernels. Here $\beta$ controls the amount of compression in the representation $T$ of $X$: small $\beta$ implies more compression (sacrificing informativeness), while larger $\beta$ pushes towards finer representation granularity that favor informativeness. Varying $\beta\in[0,+\infty)$ regulates the tradeoff between informativeness and compression.

For any $\beta\in[0,+\infty)$, conditions for a stationary point of $\cL_\beta(P_{T|X})$ can be expressed via the following self-consistent equations~\cite{tishby_IB1999}. Namely, a stationary point $P^{(\beta)}_{T|X}$ must satisfy:
\begin{subequations}
\begin{align}
P^{(\beta)}_T(t)&=\int_{\cX} P^{(\beta)}_{T|X}(t|x)\dd P_X(x)\\
P^{(\beta)}_{Y|T}(y|t)&=\frac{1}{P^{(\beta)}_T(t)}\int_{\cX} P_{Y|X}(y|x)P^{(\beta)}_{T|X}(t|x)\dd P_X(x)\\
P^{(\beta)}_{T|X}(t|x) &= \frac{P^{(\beta)}_T(t)}{Z_\beta(x)} e^{-\beta \dkl\big(P_{Y|X}(\cdot|x)\big\|P^{(\beta)}_{Y|T}(\cdot|t)\big)},
\end{align}\label{EQ:fixed_point_eq}%
\end{subequations}
where $Z_\beta(x)$ is the normalization constant (partition function). If $X,\ Y$ and $T$ take values in finite sets, and $P_{X,Y}$ is known, then alternating iterations of \eqref{EQ:fixed_point_eq} locally converges to a solution, for any initial $P_{T|X}$~\cite{tishby_IB1999}. This is reminiscent of the Blahut-Arimoto algorithm for computing rate distortion functions or channel capacity~\cite{Blahut72,Arimoto72}. The discrete alternating iterations algorithms was later adapted to the Gaussian IB~\cite{tishby_Gaussian_IB2005}.\footnote{see \cite{estella2018distributedgaussian,aguerri2018distributeddiscrete} for extensions of both the discrete and the Gaussian cases to the distributed setup.} While the algorithm is infeasible for general continuous distributions, in the Gaussian case it translates into a parameter updating procedure. 

The IB curve is obtained by solving $\inf_{P_{T|X}}\cL_\beta(P_{T|X})$, for each $\beta\in[0,+\infty)$, and plotting the mutual information pair $\big(I_\beta(X;T),I_\beta(Y;T)\big)$ for an optimal $P^{(\beta)}_{T|X}$. In Section~\ref{SUBSEC:Gaussian_IB} we show the IB curve for jointly Gaussian $(X,Y)$ variables (Fig.~\ref{FIG:IB_curve}). The two-dimensional plane in which the IB curve resides, was later coined as the \emph{information plane} (IP)~\cite{tishby_DNN2015}.


\subsection{Relation to Minimal Sufficient Statistics}\label{SUBSEC:min_suff}

A elementary concept that captures the notion of `compressed relevance' is that of MSS~\cite{lehmann1955completeness}, as defined next.

\begin{definition}[Minimal Sufficient Statistic]
Let $(X,Y)\sim P_{X,Y}$. We call $T:=t(X)$, where $t$ is a deterministic function, a \textit{sufficient statistic} of $X$ for $Y$ if $Y\leftrightarrow T\leftrightarrow X$ forms a Markov chain. A sufficient statistic $T$ is \textit{minimal} if for any other sufficient statistic $S$, there exists a function $f$, such that $T=f(S)$, $P_X$-almost surely (a.s.).
\end{definition}

The need to define minimally arises because $T=X$ is trivially sufficient, and one would like to avoid such degenerate choices. However, sufficient statistics themselves are rather restrictive, in the sense that their dimension always depends on the sample size, except when the data comes from an exponential family (cf., the Pitman–Koopman–Darmois Theorem~\cite{koopman1936distributions}). It is therefore useful to consider relaxations of the MSS framework. 

Such a relaxation is given by the IB framework, which is evident by relating it to MSSs as follows. Allowing $T$ to be a stochastic (rather than deterministic) function of $X$, i.e., defined through a transition kernel $P_{T|X}$, we have that $T$ is sufficient for $Y$ if and only if $I(X;Y)=I(X;T)$~\cite{kullback1951information}. Furthermore, set of MSSs coincides with the set of solutions of 
\begin{equation}
    \inf_{P_{T|X}\in\cF} I(X;T),\label{EQ:min_suff_stat}
\end{equation}
where $\cF:=\big\{P_{T|X}: \ I(Y;T)=\sup_{Q_{T'|X}}I(Y;T')\big\}$. Here, the mutual information terms $I(X;T)$ and $I(Y;T)$ are w.r.t. $P_{X,Y}P_{T|X}$, while $I(Y;T')$ is w.r.t. $P_{X,Y}Q_{T'|X}$.

The IB framework thus relaxes the notion of MSS in two ways. First, while a MSS is defined as a deterministic function of $X$, IB solutions are randomized mappings. Such mapping can attain strictly smaller values of $\cL_\beta(P_{T|X})$ as compared to deterministic ones.\footnote{Consider, e.g., $(X,Y)$ as correlated Bernoulli random variables.} Second, in view of \eqref{EQ:min_suff_stat}, the IB framework allows for $\beta$-approximate MSSs by regulating the amount of information $T$ retains about $Y$ (see also \eqref{EQ:IB_contrained_opt}).


\subsection{Gaussian Information Bottleneck}\label{SUBSEC:Gaussian_IB}

The Gaussian IB~\cite{tishby_Gaussian_IB2005} has a closed form solution for \eqref{EQ:IB_lagrange}, which is the focus of this section. Let $X\sim \cN(\mathbf{0},\Sigma_X)$ and $Y\sim\cN(\mathbf{0},\Sigma_Y)$ be centered multivariate jointly Gaussian (column) vectors of dimensions $d_x$ and $d_y$, respectively. Their cross-covariance matrix is $\Sigma_{XY}:=\EE\big[XY^\top\big]\in\RR^{d_x\times d_y}$.\footnote{For simplicity, we assume all matrices are full rank.}

\subsubsection{\underline{Analytic solution}}

The first step towards analytically characterizing the optimal value of $\cL_\beta(P_{T|X})$ is showing that a Gaussian $T$ is optimal. Using the entropy power inequality (EPI) in a similar vein to~\cite{berger1999semi} shows that $\inf_{P_{T|X}}\cL_\beta(P_{T|X})$ is achieved by $P^{(\beta)}_{T|X}$ for which $(X,Y,T)$ are jointly Gaussian. Since $Y\leftrightarrow X\leftrightarrow T$ forms a Markov chain, we may represent $T=\mathrm{A}X + Z$, where $Z\sim\cN(\mathbf{0},\Sigma_Z)$ is independent of $(X,Y)$. Consequently, the IB optimization problem reduces to
\begin{align*}
    \inf_{P_{T|X}}\cL_\beta(P_{T|X})&=\inf_{\mathrm{A},\Sigma_Z} I(X;\mathrm{A}X+Z)-\beta I(\mathrm{A}X+Z;Y)\\
    &=\frac{1}{2}\inf_{\mathrm{A},\Sigma_Z}\ \underbrace{(1-\beta)\log\left(\frac{\big|\mathrm{A}\Sigma_X\mathrm{A}^\top+\Sigma_Z\big|}{\big|\Sigma_Z\big|}\right)+\beta\log\Big(\big|\mathrm{A}\Sigma_{X|Y}\mathrm{A}^\top+\Sigma_Z\big|\Big)}_{:=\cL_\beta^{(\mathsf{G})}(\mathrm{A},\Sigma_Z)}\numberthis,\label{EQ:GIB}
\end{align*}
where $(X\ Y)^\top\sim \cN{\small\left(
\mathbf{0}, \begin{bmatrix}
\Sigma_X & \Sigma_{XY}\\
\Sigma_{XY}^\top & \Sigma_Y
\end{bmatrix}\right)}$ and $Z\sim\cN(\mathbf{0},\Sigma_Z)$ are independent. The second equality above follows by direct computation of the mutual information terms under the specified Gaussian law.

The optimal projection $T=\mathrm{A}X+Z$ (namely, explicit structure for $\mathrm{A}$ and $\Sigma_Z$) was characterized in~\cite[Theorem~3.1]{tishby_Gaussian_IB2005}, and is restated in the sequel. Let $\Sigma_{X|Y}= \Sigma_X-\Sigma_{XY}\Sigma_Y^{-1}\Sigma_{YX}$ be the mean squared error (MSE) matrix for estimating $X$ from $Y$. Consider its normalized form $\Sigma_{X|Y}\Sigma^{-1}$ and let $(\mathbf{v}_i)_{i=1}^k$, $1\leq k\leq d_x$ be the left eigenvectors of $\Sigma_{X|Y}\Sigma^{-1}$ with eigenvalues $(\lambda_i)_{i=1}^{k}$. We assume $(\lambda_i)_{i=1}^{k}$ are sorted in ascending order (with $(\mathbf{v}_i)_{i=1}^{k}$ ordered accordingly). For each $i=1,\ldots,k$, define $\beta^\star_i=\frac{1}{1-\lambda_i}$ and set $\alpha_i(\beta):=\frac{\beta(1-\lambda_i)-1}{\lambda_ir_i}$, where $r_i=\mathbf{v}_i^\top\Sigma_X \mathbf{v}_i$.

The tradeoff parameter $\beta\in[0,\infty)$ defines the optimal $\mathrm{A}$ matrix through its relation to the critical $(\beta^\star_i)_{i=1}^k$ values. Fix $i=1,\ldots,k+1$, and let $\beta\in[\beta^\star_{i-1},\beta^\star_i)$, where $\beta^\star_0=0$ and $\beta^\star_{k+1}=\infty$. For this $\beta$, define
\begin{equation}
\mathrm{A}_i(\beta)= \Big[\alpha_1(\beta)\mathbf{v}_1\ ,\dots,\ \alpha_i(\beta)\mathbf{v}_i\ ,\ \mathbf{0}\ ,\dots,\ \mathbf{0}\Big]^\top\in\RR^{d_x\times d_x}.
\end{equation}
Thus, $A_i(\beta)$ has $\alpha_1(\beta)\mathbf{v}_1^\top,\ldots,\alpha_i(\beta)\mathbf{v}_i^\top$ as its first $i$ rows, followed by $d_x-i$ rows of all zeros. 

\begin{theorem}[Gaussian IB~\cite{tishby_Gaussian_IB2005}]\label{TM:GIB}
Fix $\beta\in[0,\infty)$. Choosing $\Sigma_Z=\mathrm{I}_{d_x}$ and 
\begin{equation}
    \mathrm{A}(\beta):=\begin{cases}
    \mathrm{A}_0(\beta)\ ,\quad \beta\in[0,\beta_1^\star)\\
    \mathrm{A}_1(\beta)\ ,\quad \beta\in[\beta_1^\star,\beta_2^\star)\\
    \quad\quad\quad\ \vdots\\
    \mathrm{A}_k(\beta)\ ,\quad \beta\in[\beta_k^\star,\beta_{k+1}^\star)
    \end{cases},
\end{equation}
where $\mathrm{A}_0(\beta)$ is the all-zero matrix, achieves the infimum in \eqref{EQ:GIB}. 
\end{theorem}

The proof of Theorem~\ref{TM:GIB} is not difficult. First one shows that $\Sigma_Z=\mathrm{I}_{d_x}$ is optimal by a standard whitening argument. This step assumes $\Sigma_Z$ is non-singular, which does not lose generality. Indeed, if $\Sigma_Z$ is low rank then $\cL_\beta^{(\mathsf{G})}(\mathrm{A},\Sigma_Z)=\infty$. The result then follows by differentiating $\cL_\beta^{(\mathsf{G})}(\mathrm{A},\mathrm{I}_{d_x})$ with respect to $\mathrm{A}$ and solving analytically.

In light of Theorem~\ref{TM:GIB}, the optimal representation $T_\beta^{\star}=\mathrm{A}(\beta)X+Z$, is a noisy linear projection to eigenvectors of the normalized correlation matrix $\Sigma_{X|Y}\Sigma_X^{-1}$. These eigenvectors coincide with the well-known CCA basis~\cite{hotelling1935most}. However, in the Gaussian IB, $\beta$ determines the dimensionality of the projection (i.e., how many CCA basis vectors are used). As $\beta$ grows in a continuous  manner, the dimensionality of $T_\beta^\star$ increases discontinuously, with jumps at the critical $\beta^\star_i$ points. Larger $\beta$ implies higher-dimensional projections, while for $\beta\leq \beta_1^\star$, we have that $T_\beta^{\star}=\mathrm{I}_{d_x}$ comprises only noise. The Gaussian IB thus serves as a complexity measure with $\beta$-tunable projection rank. 
\begin{figure}[!t]
	\begin{center}
	        \includegraphics[scale = .5]{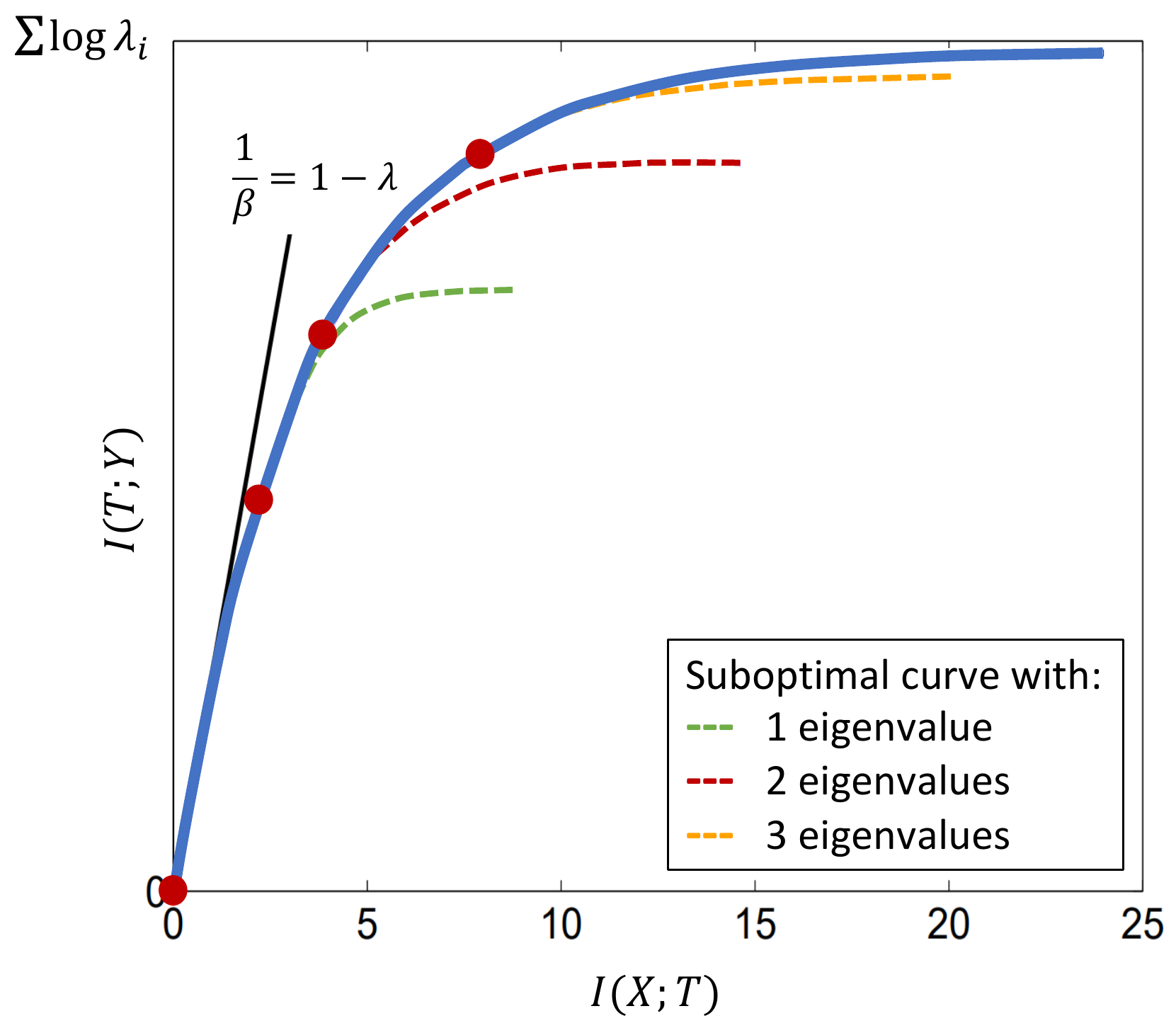}
	        \captionsetup{labelsep=none}
 	        \caption{\ [Adapted from Fig. 3 of~\cite{tishby_Gaussian_IB2005}]: The Gaussian IB curve (computed with four eigenvalues $\lambda_1=0.1$, $\lambda_2=0.5$, $\lambda_3=0.7$, $\lambda_4=0.9$ and $\beta\in[0,10^3)$) is shown in blue. Suboptimal curves that use only the first 1, 2, or 3 eigenvalues are illustrated by the dashed green, red and orange lines, respectively. Mutual information pairs at the critical values $\beta^\star_i=\frac{1}{1-\lambda_i}$, $i=1,2,3,4$, are marked by red circles. 
 	         The slope of the curve at each point is the corresponding $1/\beta$. The tangent at zero, whose slope is $1-\lambda_1$, is shown in black.}\label{FIG:IB_curve}
	\end{center}
\end{figure}

\vspace{2mm}
\subsubsection{\underline{Information bottleneck curve}} The analytic solution to the Gaussian IB problem enables plotting the corresponding IB curve. Adapted from~\cite[Fig. 3]{tishby_Gaussian_IB2005}, Fig.~\ref{FIG:IB_curve} illustrates this curve. Before discussing it we explain how it is computed. This is done by substituting the optimal projection $\mathrm{A}(\beta)$ and $\Sigma_Z=\mathrm{I}_{d_x}$ into the mutual information pair of interest. Upon simplifying, for any $\beta\in[0,\infty)$, one obtains
\begin{subequations}
\begin{align}
    I(X;T_\beta^{\star})&=\frac{1}{2}\sum_{i=1}^{n_\beta}\log\left((\beta-1)\frac{1-\lambda_i}{\lambda_i}\right)\label{EQ:pre_IB_curve1}\\
    I(T_\beta^{\star};Y)&=I(X;T_\beta)-\frac{1}{2}\sum_{i=1}^{n_\beta}\log\big(\beta(1-\lambda_i)\big),\label{EQ:pre_IB_curve2}
\end{align}\label{EQ:pre_IB_curve}%
\end{subequations}
where $n_\beta$ is the maximal index $i$ such that $\beta\geq\frac{1}{1-\lambda_i}$. Define the function
\begin{equation}
    \mathsf{F}_\mathsf{IB}(x):=x-\frac{n_{\beta(x)}}{2}\log\left(\prod_{i=1}^{n_{\beta(x)}}(1-\lambda_i)^{\frac{1}{n_{\beta(x)}}}+e^{\frac{2x}{n_{\beta(x)}}}\prod_{i=1}^{n_{\beta(x)}}\lambda_i^{\frac{1}{n_{\beta(x)}}}\right),\label{EQ:IB_curve}%
\end{equation}
where $\beta(x)$ is given by isolating $\beta$ from \eqref{EQ:pre_IB_curve1} as a function of $I(X;T_\beta^{\star})=x$. Rearranging \eqref{EQ:pre_IB_curve} one can show that $\mathsf{F}_\mathsf{IB}$ assigns each $I(X;T_\beta^{\star})$ with its corresponding $I(T_\beta^{\star};Y)$, i.e., $\mathsf{F}_\mathsf{IB}\big(I(X;T_\beta^{\star})\big)=I(T_\beta^{\star};Y)$, for all $\beta\in[0,\infty)$ (see~\cite[Section 5]{tishby_Gaussian_IB2005}).

The Gaussian IB curve, as shown in blue in Fig. \ref{FIG:IB_curve}, is computed using \eqref{EQ:IB_curve}. An interesting property of this curve is that while $\big(I(X;T_\beta^{\star}),I(T_\beta^{\star};Y)\big)$ changes continuously with $\beta$, the dimensionality of $T_\beta^{\star}$ (or, equivalently, the number of eigenvectors used in $\mathrm{A}(\beta)$) increases discontinuously at the critical points $\beta_i^\star$, $i=1,\ldots,k$, and is fixed between them. Restricting the dimension results in a suboptimal curve, that coincides with the optimal one up to a critical $\beta$ value and deviates from it afterwards. Some suboptimal curves are shown in Fig.~\ref{FIG:IB_curve} by the dashed, horizontally saturating segments. The critical values of $\big(I(X;T_\beta^{\star}),I(T_\beta^{\star};Y)\big)$ after which the suboptimal curve deviate from the optimal one are marked with red circles. Notably, the optimal curve moves from one analytic segment to another in a smooth manner. Furthermore, one readily verifies that $\mathsf{F}_\mathsf{IB}$ is a concave function with slope tends to 0 as $\beta\to\infty$. This reflects the law of diminishing returns: encoding more information about $X$ in $T$ (higher $I(T;X)$) yields smaller increments in $I(T;Y)$.

\section{Remote Source Coding}\label{SEC:remote_sc}

The IB framework is closely related to the classic information-theoretic problem of RSC. RSC dates back to the 1962 paper by Dobrushin and Tsybakov~\cite{dobrushin1962information} (see also~\cite{wolf1970transmission} for the additive noise case). As subsequently shown, the RSC rate-distortion function under logarithmic loss effectively coincides with~\ref{EQ:IB_contrained_opt}. We start by setting up the operational problem.


\subsection{Operational Setup}

Consider a source sequence $Y^n:=(Y_1,\ldots,Y_n)$ of $n$ independent and identically distributed (i.i.d.) copies of a random variable $Y\sim P_Y\in\cP(\cY)$. An encoder observes the source $Y^n$ through a memoryless noisy channel $P_{X|Y}$. Namely, $P_{X|Y}$ stochastically maps each $Y_i$ to an $\cX$-valued random variable $X_i$, where $i=1,\ldots,n$. Denoting $P_{X,Y}=P_Y P_{X|Y}$, the pair $(X^n,Y^n)$ is distributed according to its $n$-fold product measure $P_{X,Y}^{\otimes n}$. 

\begin{figure}[!t]
	\begin{center}
	        \includegraphics[scale = .6]{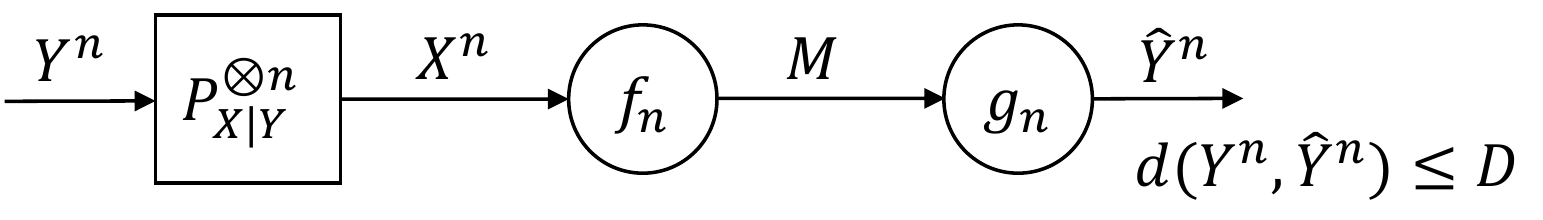}
	        \caption{Remote source coding operational setup.}\label{FIG:RSC}
	\end{center}
\end{figure}

The sequence $X^n:=(X_1,\ldots,X_n)$ is encoded through $f_n:\cX^n\to \cM_n$, where $|\cM_n|<\infty$, producing the representation $M:=f_n(X^n)\in\cM_n$. The goal is to reproduce $Y^n$ from $M$ via a decoder $g_n:\cM_n\to\hat{\cY}^n$, where $\hat{\cY}^n$ is called the reproduction space, subject to a distortion constraint.\footnote{Rate distortion theory~\cite[Section 10]{CovThom06} often considers $\hat{\cY}=\cY$, but this is not the case under logarithmic loss distortion, as described below.} The system is illustrated in Fig.~\ref{FIG:RSC}

We adopt the logarithmic loss as the distortion measure. To set it up, let the reproduction alphabet be  $\hat{\cY}=\cP(\cY)$, i.e., the set of probability measures on $\cY$. Thus, given a reproduction sequence $\hat{y}^n\in\hat{\cY}$, each $\hat{y}_i$, $i=1,\ldots,n$, is a probability measure on $\cY$. This corresponds to a `soft' estimates of the source sequence. The symbol-wise logarithmic loss distortion measure is
\begin{subequations}
\begin{equation}
d(y,\hat{y}):= \log\left(\frac{1}{\hat{y}(y)}\right),
\end{equation}
which gives rise to
\begin{equation}
d(y^n,\hat{y}^n):=\frac{1}{n}\sum_{i=1}^n \log\left(\frac{1}{\hat{y}_i(y_i)}\right)
\end{equation}
\end{subequations}
as the distortion measure between the source $y^n\in\cY^n$ and its reproduction $\hat{y}^n\in\hat{\cY}$.

An RSC block code of length $n$ comprises a pair of encoder-decoder functions $(f_n,g_n)$. A rate-distortion pair $(R,D)$ is \textit{achievable} if for every $\epsilon>0$ there exists a large enough blocklength $n$ and a code $(f_n,g_n)$, such that
\begin{equation}
\frac{1}{n}\log|\cM_n|< R+\epsilon\qquad ; \qquad \EE\big[ d(Y^n,\hat{Y}^n)\big]< D+\epsilon,
\end{equation}
where $\hat{Y}^n=g_n\big(f_n(X^n)\big)$ and $(X^n,Y^n)\sim P_{X,Y}^{\otimes n}$. The \textit{rate-distortion function} $R(D)$ is the infimum of rates $R$ such that $(R,D)$ is achievable for a given distortion $D$~\cite[Section 10.2]{CovThom06}. 

On the surface, RSC appears to be a new type of (lossy) compression problem, but it turns out to be a special case of it~\cite[Section 3.5]{berger1971rate_distortion}). Indeed, let us introduce another distortion metric $\tilde d(x, \hat y) := \EE[d(Y, \hat y)|X=x]$ and its $n$-fold extension $d(x^n, \hat y^n) = {1\over n} \sum_{i=1}^n \tilde
d(x_i, \hat y_i)$. One readily sees that $\EE\big[d(Y^n, \hat Y^n)\big] = \EE\big[\tilde d(X^n, \hat Y^n)\big]$. Thus, the standard rate-distortion theory implies that (subject to technical conditions, cf.~\cite[Chapter 26]{PolyWu-LecNotes})
\begin{equation}\label{eq:rsc_rd}
	R(D) = \inf_{P_{\hat{Y}|X}:\ \EE[\tilde d(X, \hat Y)] \le D} I(X; \hat Y).
\end{equation}

\subsection{From Rate-Distortion Function to Information Bottleneck}

Adapted to logarithmic loss, \eqref{eq:rsc_rd} can be further simplified. Indeed, consider an arbitrary distribution $P_{X,\hat Y}$ and extend it to $P_{Y,X,\hat Y}=P_{X,\hat Y} P_{Y|X}$, so that $Y\leftrightarrow X\leftrightarrow \hat Y$ forms a Markov chain. Consider any $\hat y$ and denote by $\hat y_0$ the following distribution: 
$$ \hat y_0(\cdot) := P_{Y|\hat Y=\hat y}\,.$$
From convexity, we can easily see that $\EE\big[\tilde d(X,\hat y)\big|Y=\hat y\big] \le \EE\big[\tilde d(X, \hat y_0)\big|Y=\hat y\big]$.
Therefore, any distortion minimizing $P_{X,\hat Y}$ should satisfy the condition
	\begin{equation}\label{eq:rd_rsc2}
		\hat y = P_{Y|\hat Y=\hat y}, \qquad \mbox{$P_{\hat Y}$-a.s.}\,.
\end{equation}	
In other words, we have $\EE\big[d(Y,\hat Y)\big] = \EE\big[\tilde d(X,\hat Y)\big] \ge
H(Y|\hat Y)$ with equality holding whenever condition \eqref{eq:rd_rsc2} is satisfied. Relabeling $\hat Y$ as $T$ we obtain
\begin{equation}
    R(D)=\inf_{P_{T|X}:\ H(Y|T)\leq D}I(X;T),\label{EQ:RSC_RD}
\end{equation}
where the underlying joint distribution is $P_{X,Y}P_{T|X}$, i.e., so that $Y\leftrightarrow X\leftrightarrow T$ forms a Markov chain. 

Yet another way to express IB and $R(D)$ is the following. Let $\Pi_1$ be a random variable valued in $\mathcal{P}(\cY)$ --- the space of probability measures on $\cY$. Namely, $\Pi_1 = P_{Y|X=x_0}$ with probability $P_X(x_0)$ (and, more generally, $\Pi_1 = P_{Y|X=X}$ for a regular branch of conditional law $P_{Y|X}$). Then, we have 
\begin{equation}
    R(D) = \inf_{P_{\Pi_2|\Pi_1}}\big\{I(\Pi_1; \Pi_2):\, \EE[\dkl(\Pi_1 \| \Pi_2)] \le D - H(Y|X)\big\}\,,
\end{equation}
where $\dkl$ is the Kullback-Leibler (KL) divergence and the minimization is over all conditional distributions of $\Pi_2 \in \mathcal{P}(\cY)$ given $\Pi_1$.

An interesting way to show the achievability of $R(D)$ in the RSC problem is via Berger-Tung multiterminal source coding (MSC) inner bound~\cite{Berger78,multi_Tung1978}. RSC is a special case of a 2-user MSC obtained by nullifying the communication rate of one source and driving to infinity the distortion constraint on the other. Alternatively, one may employ an explicit coding scheme: first quantize (encode) the observation $X^n$ to a codeword $T^n$, such that the empirical distribution of $(X^n,T^n)$ is a good approximation of $P_XP_{T|X}$, and then communicate this quantization to the decoder via Slepian-Wolf coding~\cite{Slepian_wolf_73_source_coding}. The reader is referred to~\cite[Section IV-B]{courtade2013multiterminal} for a detailed treatment of MSC under logarithmic loss.

Setting $D=H(Y)-\alpha$ in \eqref{EQ:RSC_RD} one recovers the IB problem \eqref{EQ:IB_contrained_opt}. Thus, RSC with logarithmic loss can be viewed as the operational setup whose solution is given by the IB optimization problem. This provides an operational interpretation to the ad hoc definition of the IB framework as presented in Section~\ref{SEC:IB}. Additional connections between the IB problem and other information-theoretic setups can be found in the recent survey \cite{zaidi2020tutorial}.


\section{Information Bottleneck in Machine Learning}\label{SEC:IB_DL}

The IB framework had impact on both theory and practice of ML. While its first applications in the field was for clustering ~\cite{slonim2000agglomerative}, more recently it is often explored as a learning objective for deep models. This was concurrently done in~\cite{alemi2017deep,higgins2017beta,achille2018information} by optimizing the IB Lagrangian \eqref{EQ:IB_lagrange} via a variational bound compatible with gradient-based methods. Applications to classification and generative modeling were explored in~\cite{alemi2017deep,achille2018information} and~\cite{higgins2017beta}, respectively. A common theme in~\cite{achille2018information} and~\cite{higgins2017beta} was that the IB objective promotes disentanglement of representations.

From a theoretical standpoint, the IB gave rise to a bold information-theoretic DL paradigm~\cite{tishby_DNN2015,DNNs_Tishby2017}. It was argued that, even when the IB objective in not explicitly optimized, DNNs trained with cross-entropy loss and SGD inherently solve the IB compression-prediction trade-off. This `IB theory for DL' attracted significant attention, culminating in many follow-up works that tested the proclaimed narrative and its accompanying empirical observations. This section surveys the practical and theoretical roles of IB in DL, focusing on classification tasks.


\subsection{Information Bottleneck as Optimization Objective}\label{SUBSEC:IB_practice}

\vspace{2mm}
\subsubsection{\underline{Variational approximation and efficient optimization}}\label{SUBSUBSEC:VIB} In~\cite{alemi2017deep}, a variational approximation to the IB objective \eqref{EQ:IB_lagrange} was proposed. The approximation parametrized the objective using a DNN and proposed an efficient training algorithm. As opposed to classic models, the obtained system is stochastic in the sense that it maps inputs to internal representations via randomized mappings. Empirical tests of the VIB system showed that it generalized better and was more robust to adversarial examples compared to competing methods.

\textbf{IB objective.} Given a DNN, regard its $\ell\textsuperscript{th}$ internal representation $T_\ell$, abbreviated as $T$, as a randomized mapping operating on the input feature $X$; the corresponding label is $Y$. Denote by $\cX$, $\cY$ and $\cT$ the sets in which $X$, $Y$ and $T$ take values. The encoding of $X$ into $T$ is defined through a conditional probability distribution, which the VIB parametrizes as $P_{T|X}^{(\theta)}$, $\theta\in\Theta$. Together with $P_{X,Y}$, $P_{T|X}
^{(\theta)}$ defines the joint distribution of $\big(X,Y,T^{(\theta)})\sim P_{X,Y,T}^{(\theta)}:=P_{X,Y}P_{T|X}^{(\theta)}$. With respect to this distribution, one may consider the optimization objective
\begin{equation}
    \cL_\beta^{\mathsf{(VIB)}}(\theta):= \max_{\theta\in\Theta}I\big(T^{(\theta)}_\ell;Y\big)-\beta I\big(X;T^{(\theta)}_\ell\big).\label{EQ:VIB}
\end{equation}
In accordance to the original IB problem~\cite{tishby_IB1999}, the goal here is to learn representations that are maximally informative about the label $Y$, subject to a compression requirement on $X$. However, since the data distribution $P_{X,Y}$ is unknown and (even knowing it) direct optimization of \eqref{EQ:VIB} is intractable --- a further approximation is needed. 

\textbf{Variational approximation.} To overcome intractability of \eqref{EQ:VIB}, the authors of~\cite{alemi2017deep} lower bound it by a form that is easily optimized via gradient-based methods. Using elementary properties of mutual information, entropy and KL divergence, \eqref{EQ:VIB} is lower bounded by
\begin{equation}
    \EE_{P^{(\theta)}_{Y,T}}\left[\log Q^{(\phi)}_{Y|T}\big(Y\big|T^{(\theta)}\big)\right]-\beta\dkl\big(P^{(\theta)}_{T|X}\big\|P_{T}^{(\theta)}\mspace{4mu}\big|P_X\big),\label{EQ:VIB_LB}
\end{equation}
where $Q^{(\phi)}_{Y|T}$ is a conditional distribution from $\cT$ to $\cY$ parametrized by a NN with parameters $\phi\in\Phi$. This distribution plays the role of a variational approximation of the decoder $P^{(\theta)}_{Y|T}$. Another difficulty is that the marginal $P_{T}^{(\theta)}(\cdot)=\int P^{(\theta)}_{T|X}(\cdot|x)\dd P_X(x)$, which is defined by $P^{(\theta)}_{T|X}$ and $P_X$, it is typically intractable. To circumvent this, one may replace it with  some reference measure $R_{T}\in\cP(\cT)$, thus further lower bounding \eqref{EQ:VIB_LB}. This is the strategy that was employed in~\cite{alemi2017deep}.


Using the reparametrization trick from~\cite{kingma2013auto} and replacing $P_{X,Y}$ in \eqref{EQ:VIB_LB} with its empirical proxy $P_n:=\frac{1}{n}\sum_{i=1}^n\delta_{(x_i,y_i)}$, where $\delta_x$ is the Dirac measure centered at $x$ and $\cD_n:=\big\{(x_i,y_i)\big\}_{i=1}^n$ is the dataset, we arrive at the empirical loss function
\begin{equation}
    \hat{\cL}_{\beta}^{(\mathsf{VIB})}(\theta,\phi,\cD_n):=\frac{1}{n}\sum_{i=1}^n\EE\Big[-\log Q^{(\phi)}_{Y|T}\big(y_n\big|f(x_n,Z)\big)\Big]+\beta\dkl\Big(P^{(\theta)}_{T|X}(\cdot|x_n)\Big\|R_{T}(\cdot)\Big).\label{EQ:VIB_ERM}
\end{equation}
Here $Z$ is an auxiliary noise variable (see~\cite{kingma2013auto}) and the expectation is w.r.t. its law. The first term is the average cross-entropy loss, which is commonly used in DL. The second term serves as a regularizer that penalizes the dependence of $X$ on $T$, thus encouraging compression. The empirical estimator in \eqref{EQ:VIB_ERM} of the variational lower bound is differentiable and easily optimized via standard stochastic gradient-based methods. The obtained gradient is an unbiased estimate of the true gradient.

\begin{figure}[!t]
	\begin{center}
	        \subfloat[]{\includegraphics[scale = .5]{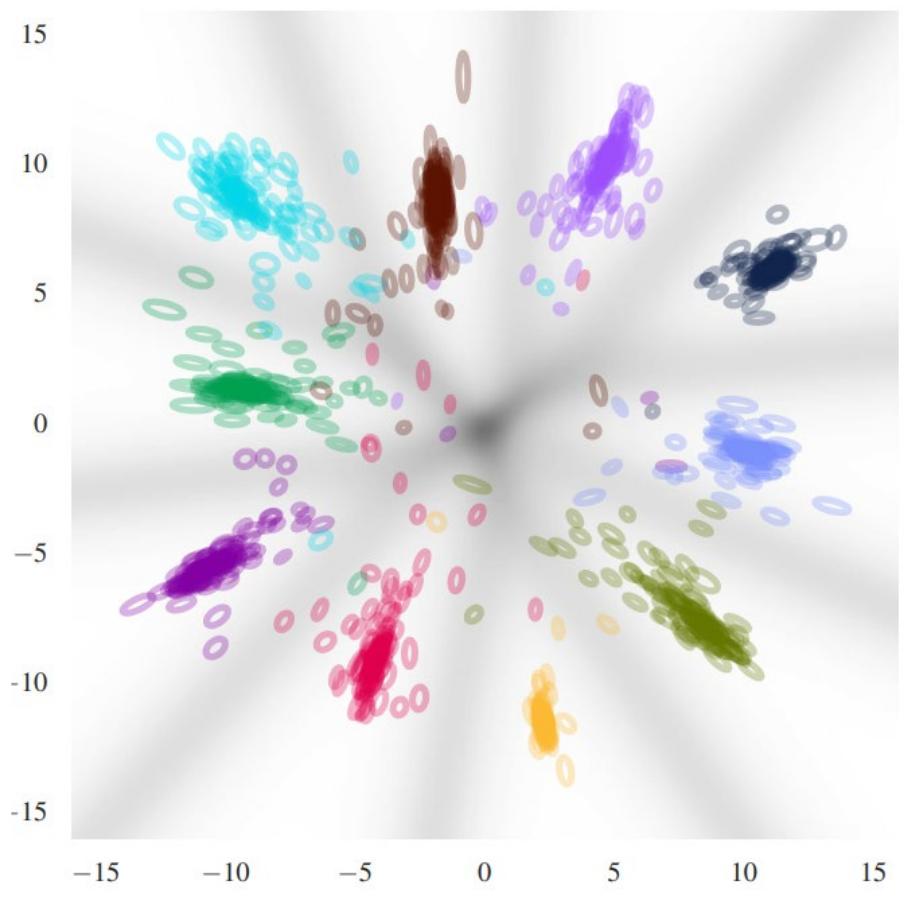}}\ \ \ \ \ 
	        \subfloat[]{\includegraphics[scale = .5]{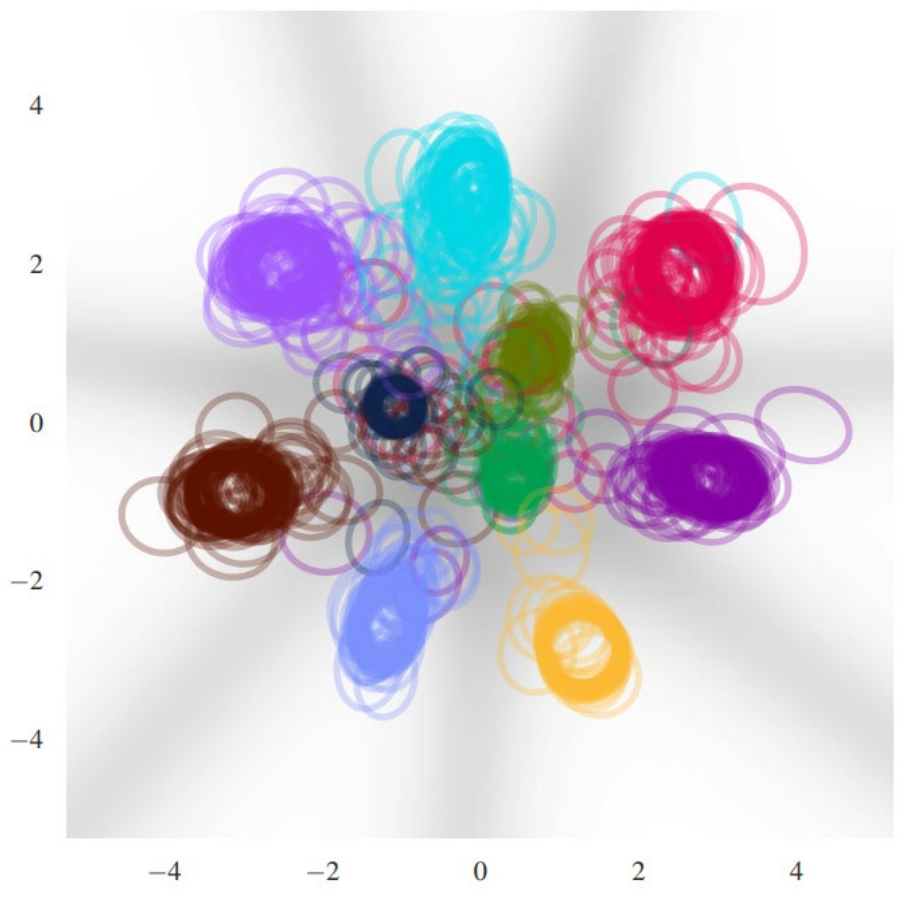}}\ \ \ \ \ 
	        \subfloat[]{\includegraphics[scale = .5]{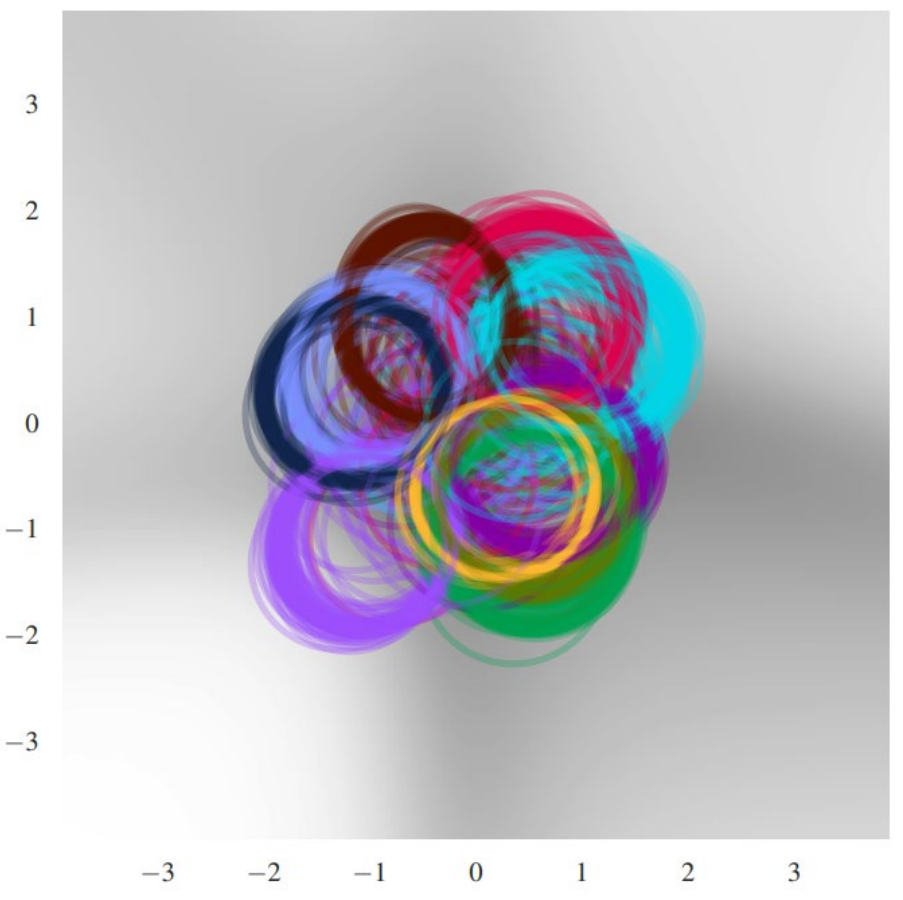}}
	        \captionsetup{labelsep=none}
 	        \caption{\ [Fig. 2 from~\cite{alemi2017deep}]: 2-dimensional VIB embedding of $10^3$ MNIST images for: (a) $\beta=10^{-3}$ ; (b) $\beta=10^{-1}$; and (c) $\beta=1$. The ellipses in each figure illustrate $95\%$ confidence intervals of the Gaussian encoding $P^{(\theta)}_{T|X}(\cdot|x)=\cN\big(\mu_\theta(x),\Sigma_\theta(x)\big)$, for $10^3$ input MNIST images $x$. The larger $\beta$ is, the more compressed representations become. This is expressed in larger covariance matrices for the encoder. The background grayscale illustrates $H\left(Q^{(\phi)}_{Y|T}(\cdot|t)\right)$, for each $t\in\RR^2$, which measures classification uncertainty at a given point.}\label{FIG:VIB}
	\end{center}
\end{figure}

\textbf{Empirical study.} For their experiments, the authors of~\cite{alemi2017deep} set the encoder distribution to $P^{(\theta)}_{T|X}(\cdot|x)=\cN\big(\mu_\theta(x),\Sigma_\theta(x)\big)$, with mean and convariance matrix parametrized by a DNN. The variational decoder $Q^{(\phi)}_{Y|T}$ was set to a logistic regression function, while $R_{T}$ was chosen as a (fixed) standard Gaussian distribution. The performance of VIB was tested on the MNIST and ImageNet datasets; we focus on the MNIST results herein. First, it was shown that a VIB classifier outperforms a multi-layer perceptron (MLP) fitted using (penalized) maximum likelihood estimation (MLE). Considered penalization techniques include dropout~\cite{Dropout2014}, confidence penalty, and label smoothing~\cite{pereyra2017regularizing} (see~\cite[Table 1]{alemi2017deep} for accuracy results).

To illustrate the operation of VIB, a 2-dimensional embedding of internal representations is examined for different $\beta$ values. The results are shown in Fig.~\ref{FIG:VIB} (reprinted from~\cite[Fig. 2]{alemi2017deep}). The posteriors $P_{T|X}^{(\theta)}$ are represented as Gaussian ellipses (representing the $95\%$ confidence region) for $10^3$ images from the test set. Colors designate true class labels. The grayscale shade in the background corresponds to the entropy of the variational classifier $Q_{Y|T}^{(\phi)}$ at a given point, i.e., $H\left(Q_{Y|T}^{(\phi)}(\cdot|t)\right)$, for $t\in\RR^2$. This entropy quantifies the uncertainty of the decoder regarding the class assignment to each point. Several interesting observations are: (i) as $\beta$ increases (corresponds to more compression in \eqref{EQ:VIB}), the Gaussian encoder covariances increase in relation to the distance between samples, and the classes start to overlap; (ii) beyond some critical $\beta$ value, the encoding `collapses' essentially losing class information; and (iii) there is uncertainty in class predictions, as measured by $H\left(Q_{Y|T}^{(\phi)}(\cdot|t)\right)$, in the areas between the class embeddings. This illustrates the ability of VIB to learn meaningful and interpretable representations of data, while preserving good classification performance.

While $\beta=10^{-1}$ (Fig.~\ref{FIG:VIB}(b)) has relatively large covariance matrices,~\cite{alemi2017deep} showed that this this system has reasonable classification performance ($3.44\%$ test error). This implies there is  significant in-class uncertainty about locations of internal representations, although the classes themselves are well-separated. Such encoding would make it hard to infer which input image corresponds to a given internal representation. This property leads to consider model robustness, which was demonstrated as another virtue of VIB.

To test model robustness,~\cite{alemi2017deep} employed the fast gradient sign (FGS)~\cite{goodfellow2014explaining} and $L^2$ optimization~\cite{carlini2017towards} attacks for perturbing inputs to fool the classifier. Robustness was measured as classification accuracy of adversarial examples. The (stochastic) VIB classifier showed increasing robustness to these attacks, compared to competing deterministic models which misclassified all the perturbed inputs. This is an outcome of the randomized VIB mapping from $\cX$ to $\cT$, which suppresses the ability to tailor adversarial examples that flip the classification label. Robustness is also related to compression: indeed, larger $\beta$ values correspond to more robust systems, which can withstand adversarial attacks with large perturbation norms (see~\cite[Fig. 3]{alemi2017deep}).

\vspace{2mm}
\subsubsection{\underline{IB objective for sufficiency, minimality and disentanglement}} Another perspective on the IB objective was proposed in~\cite{achille2018information}, that was published concurrently with~\cite{alemi2017deep}. This work provided an information-theoretic formulation of some desired properties of internal representations, such as sufficiency, minimality and disentanglement, and showed that IB optimization favors models that posses them. To optimize VIB,~\cite{achille2018information} presented a method that closely resembles that of~\cite{alemi2017deep}. The method reparametrizes $T^{(\theta)}$ as a product of some function of $X$ and a noise variable $Z$. The noise distribution is for the system designer to choose. Setting $Z\sim\mathsf{Bern}(p)$, recovers the popular dropout regularization, hence the authors of~\cite{achille2018information} called their method `information dropout'. The noise distribution employed in~\cite{achille2018information} is log-normal with zero mean, and variance that is a parametrized function of $X$. We omit further details due to similarity to VIB optimization (Section~\ref{SUBSUBSEC:VIB}).

\textbf{Minimality and sufficiency.} That IB solutions are approximate MSSs is inherent to the problem formulation (see Section~\ref{SUBSEC:min_suff}). The authors of~\cite{achille2018information} discussed the relation between minimality of representation and invariance to nuisance factors. Such factors affect the data, but the label is invariant to them. By penalizing redundancy of representations (i.e., enforcing small $I(X;T^{(\theta)})$) it was heuristically argued that the model's sensitive to nuisances is mitigated. Some experiments to support this claim were provided, demonstrating the ability of VIB to classify hand-written digits from the cluttered MNIST dataset~\cite{mnih2014recurrent}, or CIFAR-10 images occluded by MNIST digits. 

\textbf{Disentanglement.} Another property considered in~\cite{achille2018information} is disentanglement, which refers to weak dependence between elements of an internal representation vector. This idea was formalized using total correlation (TC):
\begin{equation}
    \dkl\left(P_T^{(\theta)}\middle\|\prod\nolimits_{j=1}^d Q_j\right)\label{EQ:TC}
\end{equation}
where $\prod_{j=1}^d Q_j$ is some product measure on the elements of the $d$-dimensional representation $T$. Adding a TC regularizer to \eqref{EQ:VIB_LB} will encourage the system to learn disentangled representation in the sense of small \eqref{EQ:TC}. If the TC regularization parameter is set equal to $\beta$ in \eqref{EQ:VIB_LB}, it trivially simplifies to 
\begin{equation*}
    \EE_{P^{(\theta)}_{Y,T}}\left[\log Q^{(\phi)}_{Y|T}\big(Y\big|T^{(\theta)}\big)\right]-\beta\dkl\left(P^{(\theta)}_{T|X}\middle\|\prod\nolimits_{j=1}^d Q_j\middle|P_X\right).
\end{equation*}
Thus, choosing $R_T\in\cP(\cT)$, in the framework of~\cite{alemi2017deep}, as a product measure is equivalent to regularizing for disentanglement. While the term `disentanglement' was not used in~\cite{alemi2017deep}, they do set $R_T$ as a product measure.

Altogether, Section~\ref{SUBSEC:IB_practice} demonstrates the practical usefulness of the IB as an optimization objective. It is easy to optimize under proper parameterization and learns representation with various desired properties. The work of~\cite{DNNs_Tishby2017} took these observation a step further. They claimed that DNNs trained with SGD and cross-entropy loss inherently solve the IB problem, even when there in no explicit reference to the IB problem in the system design. We elaborate on this theory next.


\subsection{Information Bottleneck Theory for Deep Learning}\label{SUBSEC:IB_theory}

Recently, a information-theoretic paradigm for DL based on the IB framework was proposed~\cite{tishby_DNN2015,DNNs_Tishby2017}. It claimed that DNN classifiers trained with cross-entropy loss and SGD inherently (try to) solve the IB optimization problem. Namely, the system aims to find internal representations that are maximally informative about the label $Y$, while compressing $X$, as captured by \eqref{EQ:IB_lagrange}. Coupled with an empirical case study, this perspective was leveraged to reason about the optimization dynamics in the IP, properties of SGD training, and the computational benefit of deep architectures. This section reviews these ideas as originally presented in~\cite{tishby_DNN2015,DNNs_Tishby2017}, with the exception of Section \ref{SUBSUBSEC:IB_vacuous} that discusses the incompatibility of the IB framework to deterministic DNNs. Whether the proposed theory holds is general was challenged by several follow-up works, which are expounded upon in Section~\ref{SEC:IB_DL_controversy}.


\vspace{2mm}
\subsubsection{\underline{Setup and Preliminaries}}\label{SUBSUBSEC:preliminaries} Consider a feedforward DNN with $L$ layers, operating on an input $x\in\mathbb{R}^{d_0}$ according to:
\begin{equation}
    \phi_\ell(x) = \sigma(\mathrm{A}_\ell\phi_{\ell-1}(x)+b_\ell)\,, \qquad \phi_0(x)=x\,, \qquad \ell=1,\ldots, L\,,\label{EQ:DNN_def}
\end{equation}
where $\mathrm{A}_\ell\in\mathbb{R}^{d_\ell\times d_{\ell-1}}$ and $b_\ell\in\mathbb{R}^{d_\ell}$ are, respectively, the $\ell\textsuperscript{th}$ weight matrix and bias vector, while  $\sigma:\mathbb{R}\to\mathbb{R}$ is the activation function operating on vectors element-wise. Letting $(X,Y)\sim P_{X,Y}$ be the feature-label pair, we call $T_\ell\triangleq\phi_\ell(X)$, $\ell=1,\ldots,L-1$, the $\ell\textsuperscript{th}$ internal representation. The output layer $T_L$ is a reproduction\footnote{Perhaps up to an additional soft-max operation.} of $Y$, sometimes denoted by $\hat{Y}$. 

The goal of the DNN classifier is to learn a reproduction $\phi_L(X)=\hat{Y}$ that is a good approximation of the true label $Y$. Let $\theta$ represent the network parameters (i.e., weight matrices and bias vectors) and write $\phi_L^{(\theta)}$ for $\phi_L$ to stress the dependence of the DNN's output on $\theta$. Statistical learning theory measures the quality of the reproduction by the \textit{population risk}
\begin{equation*}
    L_{P_{X,Y}}(\theta):=\EE c\big(\phi_L^{(\theta)}(X),Y\big)=\int c\big(\phi_L^{(\theta)}(x),y)\dd P_{X,Y}(x,y),
\end{equation*}
where $c$ is the cost/loss function. Since $P_{X,Y}$ is unknown, a learning algorithm cannot directly compute $L_{P_{X,Y}}(\theta)$ for a given $\theta\in\Theta$. Instead, it can
compute the empirical risk of $\theta$ on the dataset $\cD_n:=\big\{(x_i,y_i)\big\}_{i=1}^n$, which comprises $n$ i.i.d. samples from $P_{X,Y}$. The \textit{empirical risk} is given by
\begin{equation*}
    L_{\cD_n}(\theta):=\frac{1}{n}\sum_{i=1}^n c(\phi_L^{(\theta)}(x_i),y_i).
\end{equation*}
Minimizing $L_{\cD_n}(\theta)$ over $\theta$ is practically feasible, but the end goal is to attain small population risk. The gap between them is captured by the \textit{generalization error}
\begin{equation*}
    \mathsf{gen}(P_{X,Y},\theta):=\EE\big[L_{P_{X,Y}}(\theta)-L_{\cD_n}(\theta)\big],
\end{equation*}
where the expectation is w.r.t $P_{X,Y}^{\otimes n}$. The \textit{sample complexity} $n^\star(P_{X,Y},\epsilon,\delta)$ is the least number of samples $n$ needed to ensure $L_{P_{X,Y}}(\theta)-L_{\cD_n}(\theta)\leq \epsilon$ with probability at least $1-\delta$. To reason about generalization error and sample complexity, the IB theory for DL views the learning dynamics through the so-called `information plane'.


\vspace{2mm}
\subsubsection{\underline{The information plane}} 
Consider the joint distribution of the label, feature, internal representation and output random variables. By \eqref{EQ:DNN_def}, they form a Markov chain $Y\leftrightarrow X\leftrightarrow T_1\leftrightarrow T_2\leftrightarrow \ldots\leftrightarrow T_L$, i.e., their joint law factors as $P_{X,Y,T_1,T_2,\ldots,T_L}=P_{X,Y}P_{T_1|X}P_{T_2|T_1}\mspace{-4mu}\cdot\mspace{-4mu}\cdot\mspace{-4mu}\cdot\mspace{-1mu} P_{T_L|T_{L-1}}$.\footnote{Marginal distributions of this law are designated by keeping only the relevant subscripts.} Based on this Markov relation, the data processing inequality (DPI) implies
\begin{align}
    I(X;Y)&\geq I(T_1;Y)\geq I(T_2;Y)\geq\ldots\geq I(T_L;Y)\nonumber\\
    H(X)&\geq I(X;T_1)\geq I(X;T_2)\geq\ldots\geq I(X;T_L).\label{EQ:DPI}
\end{align}
illustrating how information inherently dissipates deeper in the network. The IB theory for DL centers around how each internal representation $T_\ell$ tries to balance its two associated mutual information term.

\begin{figure}[!t]
	\begin{center}
	        \includegraphics[scale = .4]{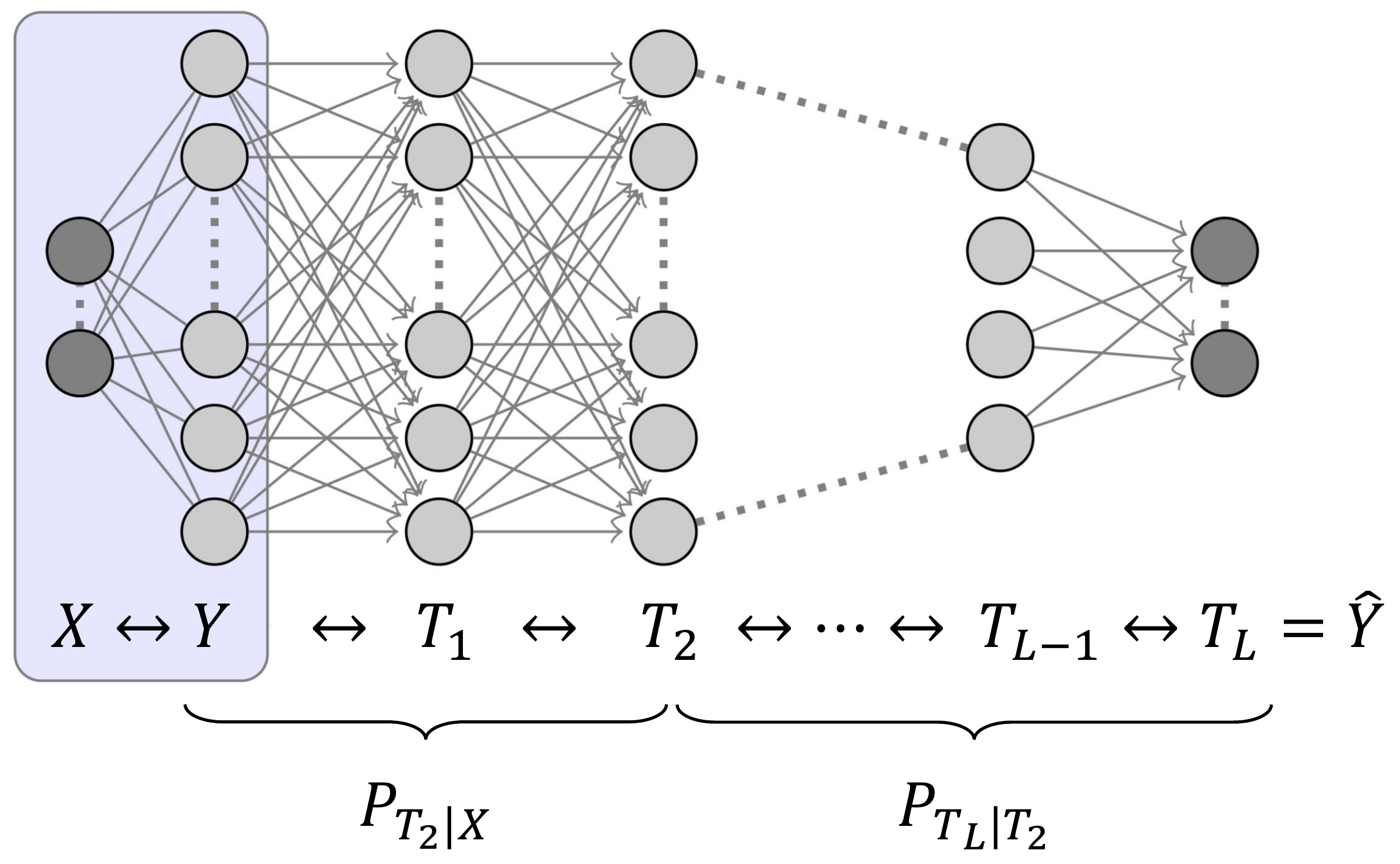}
	        \captionsetup{labelsep=none}
 	        \caption{\ [Adapted from Fig. 1 from~\cite{tishby_DNN2015}]: IB framework for a DNN classifier with $L$ layers. The label, feature, hidden representations and output form a Markov chain $Y\leftrightarrow X\leftrightarrow T_1\leftrightarrow T_2\leftrightarrow \ldots\leftrightarrow T_L$. An encoder $P_{T_\ell|X}$ and a decoder $T_{T_L|T_\ell}$ are associated with each hidden layer $T_\ell$, $\ell=1,\ldots,L-1$.}\label{FIG:IB_DNN}
	\end{center}
\end{figure}

Fix $\ell=1,\ldots,L-1$, and consider the conditional marginal distributions $P_{T_\ell|X}$ and $P_{T_L|T_\ell}$. Respectively, these distributions define an `encoder' (of $X$ into $T_\ell$) and `decoder' (of $T_\ell$ into $T_L=\hat{Y}$) for an IB problem associated with the $\ell\textsuperscript{th}$ hidden layer (see Fig.~\ref{FIG:IB_DNN}). The argument of~\cite{tishby_DNN2015} and~\cite{DNNs_Tishby2017} is that the IB framework captures the essence of learning to classify $Y$ from $X$: to reconstruct $Y$ from $X$, the latter has to go through the bottleneck $T_\ell$. This $T_\ell$ should shed information about $X$ that is redundant/irrelevant for determining $Y$, while staying maximally informative about $Y$ itself. The working assumption of~\cite{tishby_DNN2015,DNNs_Tishby2017} is that by optimizing the DNN's layers $\{T_\ell\}_{\ell=1}^L$ for that task via standard SGD, the encoder-decoder pair for each hidden layer converges to its optimal IB solution. 

With this perspective,~\cite{DNNs_Tishby2017} conducted an empirical case study of a DNN classifier trained on a synthetic task, reporting several striking findings. The study is centered around IP visualization of the learning dynamics, i.e., tracking the evolution of $\big(I(X;T_\ell),I(T_\ell;Y)\big)$ across training epochs. First, they observed that the training process comprises two consecutive phases, termed `fitting' and `compression': after a quick initial fitting phase, almost the entire training process is dedicated to compressing $X$, while staying informative about $Y$. Second, it was observed that the compression phase starts when the mean and variance of the stochastic gradient undergoes a phase transition from high to low signal-to-noise (SNR). Third,~\cite{DNNs_Tishby2017} observed that the two training phases accelerate when more layers are added to the network, which led to an argument that the benefit of deeper architectures is computational.

Central to the empirical results of~\cite{DNNs_Tishby2017} is the ability to measure $I(X;T_\ell)$ and $I(T_\ell;Y)$ in a DNN with fixed parameters. We explain next that these information measures are ill-posed in deterministic networks (i.e., networks that define a deterministic mapping from input to output). Afterwards, we describe how~\cite{DNNs_Tishby2017}  circumvented this issue by instead evaluating the mutual information terms after quantizing the internal representation vectors. We then turn to expound upon the empirical findings of~\cite{DNNs_Tishby2017} based on these quantized measurements. 




\subsubsection{\underline{Vacuous mutual information in deterministic deep networks}}\label{SUBSUBSEC:IB_vacuous}

We address theoretical caveats in applying IB-based reasoning to deterministic DNNs. A deterministic DNN is one whose output, as well as every internal representation, is a deterministic function of its input $X$. The setup described in Section~\ref{SUBSUBSEC:preliminaries}, which is standard for DNNs and the one used in~\cite{tishby_DNN2015,DNNs_Tishby2017,DNNs_ICLR2018}, adheres to the deterministic framework. In deterministic DNNs with continuous and strictly monotone nonlinearities (e.g., $\tanh$ or $\sig$) or bi-Lipschitz (e.g., $\mathrm{leaky}$-$\mathrm{ReLU}$), the mutual information $I(X;T_\ell)$ is provably either infinite (continuous $X$) or a constant that does not depend on the network parameters (discrete $X$). The behavior for $\relu$ or step activation functions is slightly different, though other issues arise, such as the IB functional being piecewise constant.

We start from continuous inputs. Elementary information-theoretic arguments show that $I(X;T_\ell)=\infty$ a.s.\footnote{With respect to, e.g., the Lebesgue measure on the parameter space of $\big\{(\mathrm{A}_i,b_i)\big\}_{i=1}^\ell$, or the entry-wise i.i.d. Gaussian measure.} if $X$ is a continuous $\RR^d$-valued random variable and nonlinearities are continuous and strictly monotone. Indeed, in this case the $d_\ell$-dimensional $T_\ell=\phi_\ell(X)$ is a.s. continuous whenever $d_\ell\leq d$, and so $I(X;T_\ell)\geq I(T_\ell;T_\ell)=\infty$ (cf.~\cite[Theorems 2.3 and 2.4]{PolyWu-LecNotes}). A more general statement was given in~\cite[Theorem 1]{amjad2019learning} and is restated next.

\begin{theorem}[Theorem 1 of~\cite{amjad2019learning}]\label{TM:MI_vacuous}
Let $X$ be a $d$-dimensional random variable, whose distribution has an absolutely continuous component with density function that is continuous on a compact subset of $\RR^d$. Assume that the activation function $\sigma$ in \eqref{EQ:DNN_def} is either bi-Lipschitz or continuously differentiable with strictly positive derivative. Then, for every $\ell=1,\ldots, L$ and almost all weight matrices $\mathrm{A}_1,\ldots,\mathrm{A}_\ell$, we have $I(X;T_\ell) =\infty$.
\end{theorem}

The proof uses the notion of correlation dimension ~\cite{csiszar1962dimension}. It shows that $X$ has a positive correlation dimension that remains positive throughout the DNN, from which the conclusion follows. This result broadens the conditions for which $I(X;T_\ell) =\infty$ beyond requiring that $T_\ell$ is continuous. This, for instance, accounts for cases when the number of neurons $d_\ell$ in $T_\ell$ exceeds the dimension of $X$. Theorem~\ref{TM:MI_vacuous} implies that for continuous features $X$, the \emph{true} mutual information $I(X;T_\ell)=\infty$, for any hidden layer $\ell=1,\ldots,L$ in the $\tanh$ network from~\cite{DNNs_Tishby2017}.

To avoid this issue, 
\cite{DNNs_Tishby2017} model $X\sim\mathsf{Unif}(\cX_n)$, where $\cX_n=\{x_i\}_{i=1}^n$. While having a discrete distribution for $X$ ensures mutual information is finite, as $I(X;T_\ell)\leq H(X)=\log n$, a different problem arises. Specifically, whenever nonlinearities are injective (e.g., strictly monotone), the map from $\cX_n$ to $\phi_\ell(\cX_n)=\{\phi_\ell(x):\ x\in\cX_n\}$ (as a mapping between discrete sets) is a.s. injective. As such, we have that $I(X;T_\ell)=H(X)=\log n$ and $I(T_\ell;Y)=I(X;Y)$, which are constants independent of the network parameters.

Both continuous and discrete degeneracies are a consequence of the deterministic DNN's ability to encode information about $X$ in arbitrarily fine variations of $T_\ell$, essentially without loss, even if deeper layers have fewer neurons. Consequently, no information about $X$ is lost when traversing the network's layers, which renders $I(X;T_\ell)$ a vacuous quantity for almost all network parameters. In such cases approximating or estimating $I(X;T_\ell)$ and $I(T_\ell;Y)$ to study DNN learning dynamics is unwarranted. Indeed, the true value (e.g., infinity or $H(X)$ for $I(X;T_\ell)$) is known and does not depend on the network.


\vspace{2mm}
\subsubsection{\underline{Mutual information measurement via quantization}}


The issues described above are circumvented in~\cite{DNNs_Tishby2017} by imposing a concrete model on $(X,Y)$ and quantizing internal representation vectors. Namely, $X$ is assumed to be uniformly distributed over the dataset $\cX_n=\{x_i\}_{i=1}^n$, while $P_{Y|X}$ is defined through a logistic regression with respect to a certain spherically symmetric real-valued function of $X$. With this model,~\cite{DNNs_Tishby2017} quantize $T_\ell$ to compute $\big(I(X;T_\ell),I(T_\ell,Y)\big)$, as described nex.

Specifically, consider a DNN with bounded nonlinearities $\sigma:\mathbb{R}\to[a,b]$. Let $\mathsf{Q}_m[T_\ell]$ be the quantized version of the $d_\ell$-dimensional random vector $T_\ell$, which dissects its support (the hypercube $[a,b]^{d_\ell}$) into $m^{d_\ell}$ equal-sized cells ($m$ in each direction). The two mutual information terms are then approximated by 
\begin{subequations}
\begin{align}
    I(X;T_\ell)&\approx I\big(X;\mathsf{Q}_m[T_\ell]\big)=H\big(\mathsf{Q}_m[T_\ell]\big)\label{EQ:bin_proxy1}\\
    I(T_\ell;Y)&\approx I\big(\mathsf{Q}_m[T_\ell];Y\big)=H\big(\mathsf{Q}_m[T_\ell]\big)-\sum_{y=0,1} P_Y(y)H\big(\mathsf{Q}_m[T_\ell]\big|Y=y\big)\label{EQ:bin_proxy2}
\end{align}\label{EQ:bin_proxy}%
\end{subequations}
where the equality in \eqref{EQ:bin_proxy1} is because $T_\ell$ (and thus $\mathsf{Q}_m[T_\ell]$) is a deterministic function of $X$, while $P_{Y}$ in \eqref{EQ:bin_proxy1} is given by $P_{Y}(1)=\frac{1}{n}\sum_{i=1}^nP_{Y|X}(1|x_i)$. Computing $H\big(\mathsf{Q}_m[T_\ell]\big)$ amounts to counting how many inputs $x_i$, $i=1,\ldots,n$, have their internal representation $\phi_\ell(x_i)$ fall inside each of the $m^{d_\ell}$ quantization cells under $\mathsf{Q}_m$. For the conditional entropy, counting is restricted to $x_i$'s whose corresponding label is $y_i=y$. 

The quantized mutual information proxies are motivated by the fact that for any random variable pair $(A,B)$,
\begin{equation}
I(A;B)=\lim_{j,k\to\infty}I\big([A]_j;[B]_k\big),\label{EQ:MI_quantization}
\end{equation}
where $[A]_j$ and $[B]_k$ are any sequences of finite quantizations of $A$
and $B$, respectively, such that the quantization errors tend to
zero as $j,k\to\infty$ (see~\cite[Section 2.3]{ElGamal2011}). Thus, at the limit of infinitely-fine quantization, the approximations in \eqref{EQ:bin_proxy} are exact. In practice,~\cite{DNNs_Tishby2017} fix the resolution $m$ in $Q_m$ and perform the computation w.r.t. this resolution. Doing so generally bounds the approximation away from the true value, resulting in a discrepancy between the computed values and the true DNN model (where activations are not quantized during training or inference). This discrepancy and its effect on the observations of~\cite{DNNs_Tishby2017} are discussed in detail in Section~\ref{SEC:IB_DL_controversy}. The remainder of the current section focuses on describing the findings of~\cite{DNNs_Tishby2017}, temporarily overlooking these subtleties.


\vspace{2mm}
\subsubsection{\underline{Information plane dynamics and two training phases}}\label{SUBSUBSEC:IP_dynamics} Using the above methodology for approximating $\big(I(X;T_\ell),I(T_\ell,Y)\big)$,~\cite{DNNs_Tishby2017} explores IP dynamics during training of a DNN classifier on a certain synthetic task. The task is binary classification of 12-dimensional inputs using a fully connected 12--10--7--5--4--3--2 architecture with $\tanh$ nonlinearities.\footnote{For the full experimental setup see~\cite[Section 3.1]{DNNs_Tishby2017}.} Training is performed via standard SGD and cross-entropy loss. The IP visualizations are produced by subsampling training epochs and computing $\big(I\big(X;\mathsf{Q}_m[T_\ell]\big),I\big(\mathsf{Q}_m[T_\ell];Y\big)\big)$, with $m=30$, w.r.t. the (fixed) DNN parameters at each epoch. To smooth out the curves, trajectories are averaged over 50 runs.

\begin{figure}[!t]
	\begin{center}
	        \includegraphics[scale = .34]{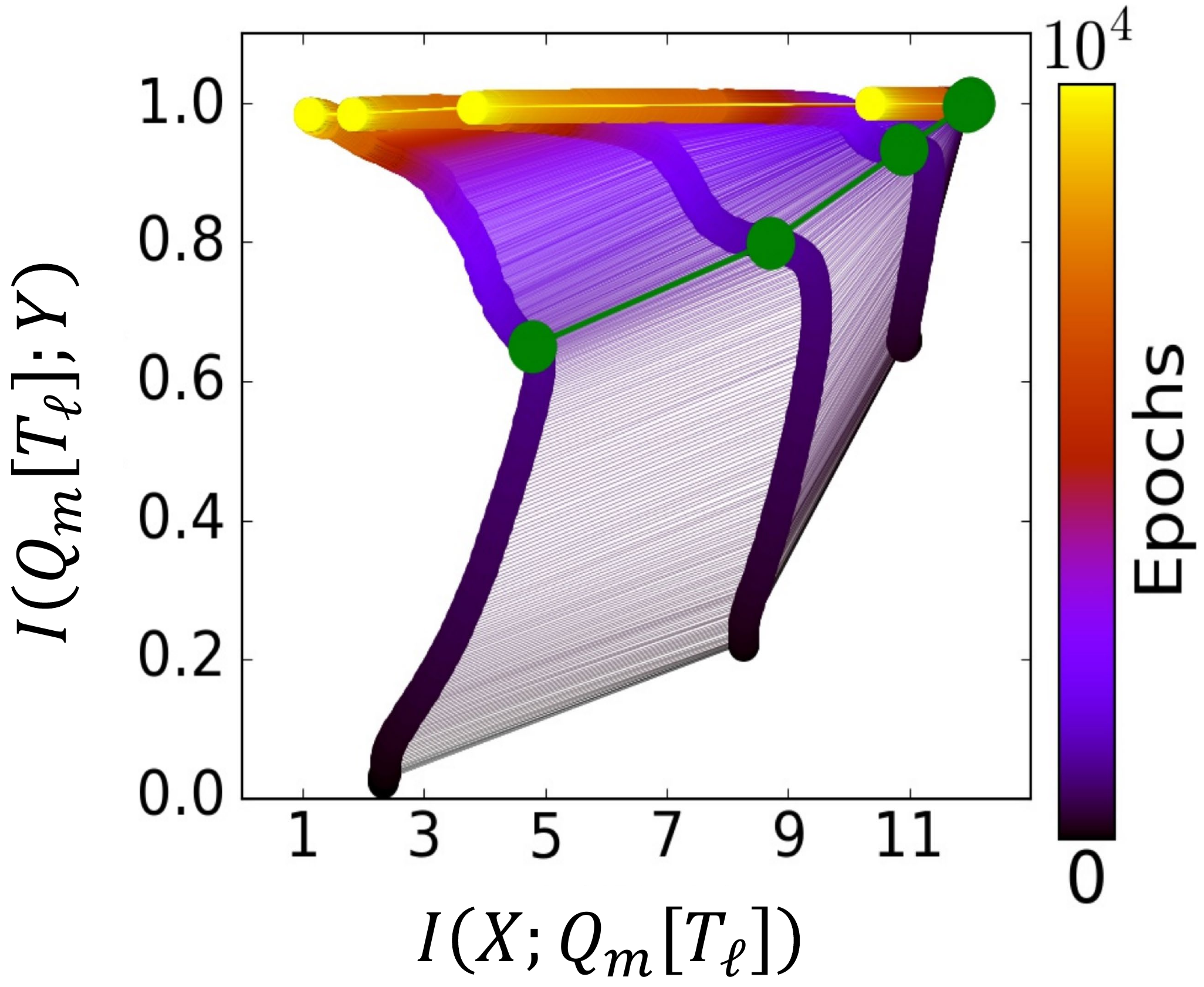}
	        \captionsetup{labelsep=none}
 	        \caption{\ [Fig. 3 from~\cite{DNNs_Tishby2017}]: IP dynamics for the fully connected 12--10--7--5--4--3--2 binary DNN classifier from~\cite{DNNs_Tishby2017}. The horizontal and vertical axis correspond to $I\big(X;\mathsf{Q}_m[T_\ell]\big)$ and $I\big(\mathsf{Q}_m[T_\ell];Y\big)$, respectively. The thick curves show $\big(I\big(X;\mathsf{Q}_m[T_\ell]\big),I\big(\mathsf{Q}_m[T_\ell];Y\big)\big)$ values across training epoch (designated by the color map) for the different hidden layers. Curves for deeper layers appear more to the left. The thin lines between the curves connect $\big(I\big(X;\mathsf{Q}_m[T_\ell]\big),I\big(\mathsf{Q}_m[T_\ell];Y\big)\big)$ values across layers at a given epoch. The two training phases of `fitting' and `compression' are clearly seen in this example, as the curve for each layer exhibits an elbow effect. The green marks correspond to the transition between the `drift' and `diffusion' phases of SGD (see Section~\ref{SUBSUBSEC:SGD_dynamics} and Fig.~\ref{FIG:sgd_phases})}\label{FIG:phases}
	\end{center}
\end{figure}

Fig.~\ref{FIG:phases} (reprinted from~\cite[Fig. 3]{DNNs_Tishby2017}), demonstrates the IP dynamics for this experiment. The thick trajectories show $\big(I\big(X;\mathsf{Q}_m[T_\ell]\big),I\big(\mathsf{Q}_m[T_\ell];Y\big)\big)$ evolution across training epochs (which are designated by the color map) for the different hidden layers of the network. Deeper layers are bound from above (in the Pareto sense) by shallower ones, in accordance to the DPI (see Eq. \eqref{EQ:DPI}). The thin lines between the IP curves connect the mutual information pair values across layers at a given epoch. These IP dynamics reveal a remarkable trend, as the trajectories of $\big(I\big(X;\mathsf{Q}_m[T_\ell]\big),I\big(\mathsf{Q}_m[T_\ell];Y\big)\big)$ exhibit two distinct phases: an increase in both $I\big(X;\mathsf{Q}_m[T_\ell]\big)$ and $I\big(\mathsf{Q}_m[T_\ell];Y\big)$ at the beginning of training, followed by along-term decrease in $I\big(X;\mathsf{Q}_m[T_\ell]\big)$ that subsumes most of the training epochs. These two phases were termed, respectively, `fitting' and `compression'. While the increase in $I\big(\mathsf{Q}_m[T_\ell];Y\big)$ during the fitting phase is expected, there was no explicit regularization to encourage compression of representations. 

Inspired by the classic IB problem (Section~\ref{SEC:IB}), the compression phase was interpreted as the network `shedding' information about $X$ that is `irrelevant' for the classification task. The authors of~\cite{DNNs_Tishby2017} then argued that the observed two-phased dynamics are inherent to DNN classifiers trained with SGD, even when the optimization method/objective have no explicit reference to the IB principle. The compression phase was further claimed to be responsible for the outstanding generalization performance of deep networks, although no rigorous connection between $I\big(X;\mathsf{Q}_m[T_\ell]\big)$ and the generalization error is currently known.


\vspace{2mm}
\subsubsection{\underline{Connection to stochastic gradient descent dynamics}}\label{SUBSUBSEC:SGD_dynamics}

\begin{figure}[!t]
	\begin{center}
	        \includegraphics[scale = .45]{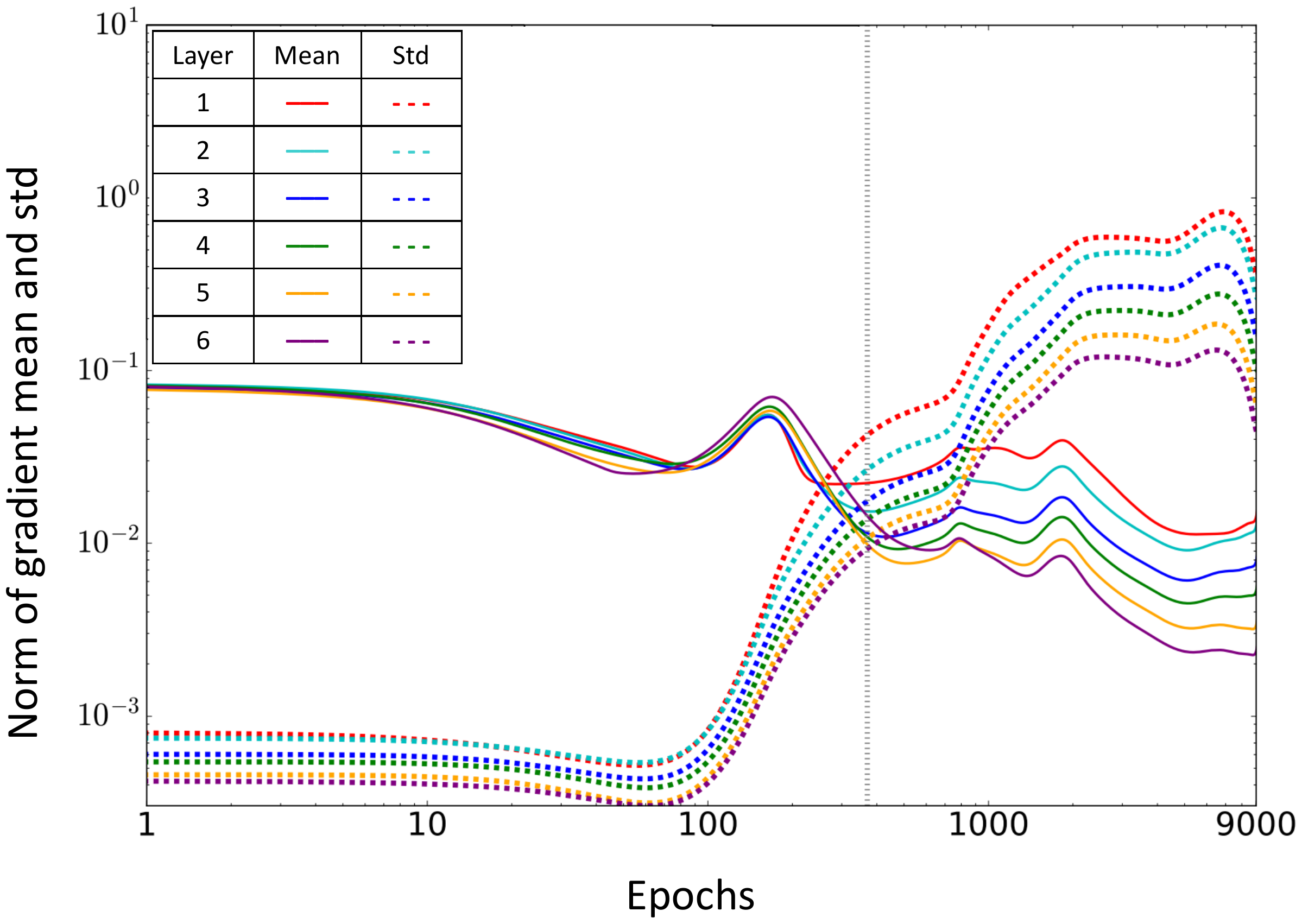}
	        \captionsetup{labelsep=none}
 	        \caption{\ [Fig. 4 from~\cite{DNNs_Tishby2017}]: Norms of gradient mean (solid lines) and standard deviation (dashed lines) for the different layers. For each layer, the values are normalized by the $L^2$ norm of the corresponding weight matrix. The grey vertical line marks the epoch when a phase transition between a high-SNR to a low-SNR occurs.}\label{FIG:sgd_phases}
	\end{center}
\end{figure}

To further understand the two IP phases of training,~\cite{DNNs_Tishby2017} compared them with SGD dynamics. For each layer $\ell=1,\ldots,L$, define
\begin{equation*}
    \mu_\ell:=\left\|\left\langle\frac{\partial c}{\partial \mathrm{A}_\ell}\right\rangle\right\|_\mathsf{F}\quad ;\quad \sigma_\ell:=\left\|\mathsf{STD}\left(\frac{\partial c}{\partial \mathrm{A}_\ell}\right)\right\|_\mathsf{F},
\end{equation*}
where $\langle\cdot\rangle$ and $\mathsf{STD}(\cdot)$ denote, respectively, the mean and the element-wise standard deviation (std) across samples within a minibatch, and $\|\cdot\|_\mathsf{F}$ is the Frobenius norm. Thus, $\mu_\ell$ and $\sigma_\ell$ capture the gradient's mean and std w.r.t. to the $\ell\textsuperscript{th}$ weight matrix. Since weights tend to grow during training, $\mu_\ell$ and $\sigma_\ell$ were normalized by $\|\mathrm{A}_\ell\|_\mathsf{F}$. The evolution of the normalized mean and std during training is displayed in Fig.~\ref{FIG:sgd_phases} (reprinted from~\cite[Fig. 4]{DNNs_Tishby2017}).  

The figure shows a clear phase transition around epoch 350, marked with the vertical grey line. At the first phase, termed `drift', the gradient mean $\mu_\ell$ is much larger than its fluctuations, as measured by $\sigma_\ell$. Casting $\mu_\ell/\sigma_\ell$ as the gradient signal-to-noise ratio (SNR) at the $\ell\textsuperscript{th}$ layer, the drift phase is characterized by high SNR. This corresponds to SGD exploring the high-dimensional loss landscape, quickly converging from the random initialization to a near (locally) optimal region. In the second phase, termed `diffusion', the gradient SNR abruptly drops. The low SNR regime, as explained in~\cite{DNNs_Tishby2017}, is a consequence of empirical error saturating and SGD being dominated by its fluctuations. This observation corresponds to the earlier work of~\cite{murata1998statistical,chee2018convergence}, where two phases of gradient descent (convergence towards a near-optimal region and oscillation in that region) were also identified and described in greater generality . Rather than `drift' and `compression', these phases are sometimes termed `transient' and `stochastic' or `search' and `convergence'.

The correspondence between the two SGD phases and the IP trajectories from Fig.~\ref{FIG:phases} was summarized in~\cite{DNNs_Tishby2017} as follows. First, the transition from `fitting' to `compression' in Fig.~\ref{FIG:phases} happens roughly at the same epoch when SGD transitions from `drift' to `diffusion' (Fig.~\ref{FIG:sgd_phases}) --- this is illustrated by the green marks in Fig.~\ref{FIG:phases}. The SGD drift phase quickly reduces empirical error, thereby increasing $I(T_\ell;Y)$ (as captured by its approximation $I\big(\mathsf{Q}_m[T_\ell];Y\big)$ in Fig.~\ref{FIG:phases}). The connection between the SGD diffusion phase and compression of $I\big(X;\mathsf{Q}_m[T_\ell]\big)$ in the IP is less clear. A heuristic argument given in~\cite{DNNs_Tishby2017} is that SGD diffusion mostly adds random noise to the weights, evolving them like Wiener processes. This diffusion-like behaviour inherently relies on the randomness in SGD (as opposed to, e.g., full gradient descent). Based on this hypothesis,~\cite{DNNs_Tishby2017} claimed that SGD diffusion can be described by the Fokker-Planck equation, subject to a small training error constraint. Together with the maximum entropy principle, this led to the conclusion that the diffusion phase maximizes the conditional entropy $H(X|T_\ell)$, or equivalently, minimizes $I(X;T_\ell)$ (approximated by $I\big(X;\mathsf{Q}_m[T_\ell]\big)$ in Fig.~\ref{FIG:phases}). We note that~\cite{DNNs_Tishby2017} presented no rigorous derivations to support this explanation.


\vspace{2mm}
\subsubsection{\underline{Computation benefit of deep architectures}} It is known that neural networks gain in representation power with the addition of layers~\cite{montufar2014number,eldan2016power,telgarsky2016benefits,poggio2017and}. The argument of~\cite{DNNs_Tishby2017} is that deep architectures also result in a computational benefit. Specifically, adding more layers speeds up both the fitting and the compression phases in the IP dynamics. 

\begin{figure}[!t]
	\begin{center}
	        \includegraphics[scale = .5]{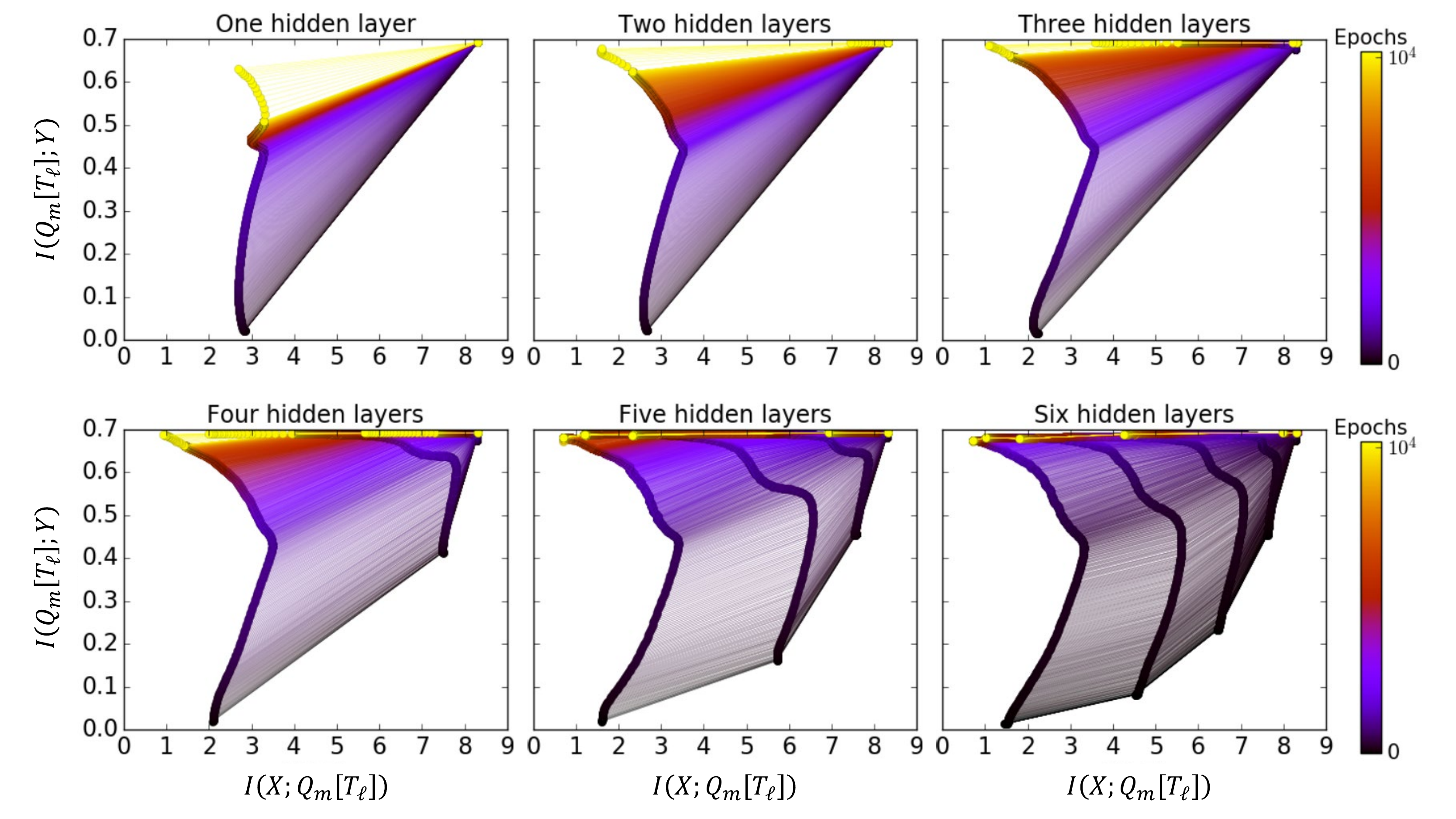}
	        \captionsetup{labelsep=none}
 	        \caption{\ [Fig. 5 from~\cite{DNNs_Tishby2017}]: IP dynamics for 6 neural networks with increasing depths. Each network has 12 input and 2 output units. The width of the hidden layers start from 12 and reduces by 2 with each added layer.}\label{FIG:IP_comp}
	\end{center}
\end{figure}

Recall the synthetic binary classification task of 12-dimensional inputs described in Section~\ref{SUBSUBSEC:IP_dynamics}. Consider 6 neural network architectures of increasing depth (1 to 6 hidden layers) trained to solve that task. The first hidden layer has 12 neurons and each succeeding one has two neurons less. For example, the 3rd architecture is a fully connected 12--12--10--8--2 networks. Fig.~\ref{FIG:IP_comp} (reprinted from~\cite[Fig. 5]{DNNs_Tishby2017}) shows the IP dynamics of these 6 architectures, each averaged over 50~runs.

The figure shows that additional layers speed up the IP dynamics. For instance, while last hidden layer of the deepest network (bottom-right subfigure) attains its maximal $I\big(\mathsf{Q}_m[T_6];Y\big)\approx 0.7$ after about 400 epochs, the shallowest network (top-left subfigure) does not reach that value throughout the entire training process. The compression of $I\big(X;\mathsf{Q}_m[T_\ell]\big)$, for $\ell=1,\ldots,L$, is also faster and more pronounced in deeper architectures. Thus, both IP phases accelerate as a result of more hidden layers. Furthermore, compression of a preceding $I\big(X;\mathsf{Q}_m[T_{\ell-1}]\big)$ seems to push $I\big(X;\mathsf{Q}_m[T_\ell]\big)$ of the next layer to a smaller value.\footnote{For the true (unquantized) $I(X;T_\ell)$ terms, $\ell=1,\ldots,L$, their reduction with larger $\ell$ values is a consequence of the DPI. The DPI, however, does not hold for the quantized mutual information proxies.} We also note that IP values of shallow layers, even in deeper architectures, stay almost unchanged throughout training. These values are roughly $\big(H(X),I(X;Y)\big)$, which correspond to $\big(I(X;T_\ell),I(T_\ell;Y)\big)$ when $T_\ell=\phi_\ell(X)$ is a bijection (as a mapping from $\cX_n$ to $\phi_\ell(\cX_n):=\{\phi_\ell(x):\ x\in\cX_n\}$). This fact will be revisited and explained in the Section~\ref{SEC:IB_DL_controversy}.

The interpretation of Fig.~\ref{FIG:IP_comp} in~\cite{DNNs_Tishby2017} (Fig. 5 therein) synonymizes `high $I\big(\mathsf{Q}_m[T_L];Y\big)$' and `good generalization'. We refrain from this wording since there is no known rigorous connection between generalization error and $I\big(\mathsf{Q}_m[T_L];Y\big)$ (or $I(T_L;Y)$ for that matter).\footnote{Some connections between other information measures and generalization error are known -- see, e.g.,~\cite{xu2017information}.} If one adopts the IP values prescribed by the IB problem as the figure of merit, then indeed, deeper architectures enhance the corresponding dynamics. However, the implications of the IB theory on how depth affects generalization error or sample complexity remains unclear.


\vspace{2mm}
\subsubsection{\underline{Concluding remarks}} Several claims from~\cite{DNNs_Tishby2017} were not discussed here in detail. Our summary focused on the empirical observations of that work. Beyond these,~\cite{DNNs_Tishby2017} included heuristic arguments about how: (i) compression of representation is causally related to the outstanding generalization performance DNNs enjoy in practice; (ii) learned hidden layers lie on (or very close to) the IB theoretical bound \eqref{EQ:IB_lagrange} for different $\beta$ values; (iii) the corresponding `encoder' and `decoder' maps satisfy the IB self-consistent equations \eqref{EQ:fixed_point_eq}; (iv) the effect of depth on the diffusion phase of SGD dynamics; and more. The reader is referred to the original paper for further details.

In sum, the IB theory for DL~\cite{tishby_DNN2015,DNNs_Tishby2017} proposed a novel perspective on DNNs and their learning dynamics. At its core, the theory aims to summarize each hidden layer into the mutual information pair $\big(I(X;T_\ell),I(T_\ell;Y)\big)$, $\ell=1,\ldots,L$, and study the systems through that lens. As the true mutual information terms degenerate in deterministic networks,~\cite{DNNs_Tishby2017} adopted the quantized versions $I\big(X;\mathsf{Q}_m[T_L]\big)$ and
$I\big(\mathsf{Q}_m[T_L];Y\big)$ as figures of merit in their stead. While doing so created a gap between the empirical study of~\cite{DNNs_Tishby2017} and the theoretical IB problem, the evolution of the quantized terms throughout training revealed remarkable empirical trends. These observations were collected to a new information-theoretic paradigm to explain DL, which inspired multiple follow-up works. The next section describes some of these works and how they corroborate or challenge claims made in~\cite{tishby_DNN2015,DNNs_Tishby2017}.


\section{Revisiting the Information Bottleneck Theory for Deep Learning}\label{SEC:IB_DL_controversy}

Since~\cite{tishby_DNN2015,DNNs_Tishby2017}, the IB problem and the IP dynamics became sources of interest in DL research. Many works followed up both on the empirical findings and theoretical reasoning in~\cite{tishby_DNN2015,DNNs_Tishby2017}; a nonexhastive list includes~\cite{Achille2018,DNNs_ICLR2018,amjad2019learning,yu2018understanding,cheng2018evaluating,ICML_Info_flow2019,cvitkovic2019minimal,wickstrom2019information,goldfeld2019convergence}. This section focuses on the empirical study conducted in~\cite{DNNs_ICLR2018}, how quantization misestimates mutual information in deterministic networks (in light of the observations from~\cite{amjad2019learning,ICML_Info_flow2019}), and the relation between compression and clustering revealed in~\cite{ICML_Info_flow2019}.


\subsection{Revisiting the Empirical Study}\label{SUBSEC:ICLR}

In~\cite{DNNs_ICLR2018}, an empirical study aiming to test the main claims of~\cite{DNNs_Tishby2017} was conducted. They focused on the two phases of training in the IP, the relation between compression of $I(X;T_\ell)$ and generalization, as well as the link between stochasticity in gradients and compression. Through a series of experiments,~\cite{DNNs_ICLR2018} produced counter examples to all these claims. We note that~\cite{DNNs_ICLR2018} employed methods similar to~\cite{DNNs_Tishby2017} for evaluating $I(X;T_\ell)$ and $I(T_\ell;Y)$. As subsequently explained in Sections~\ref{SUBSEC:misestimation} and~\ref{SUBSEC:IB_clustering}, these methods fail to capture the true mutual information values, which are vacuous in deterministic DNNs (i.e., when for fixed parameters, the DNN's output is a deterministic function of the input). Though the authors of~\cite{DNNs_ICLR2018} were aware of the this issue (see p. 5 and Appendix C therein), they adopted the methodology of~\cite{DNNs_Tishby2017} in favor of empirical comparability.\footnote{Several methods for computing IP trajectories were employed in \cite{DNNs_ICLR2018}. While they mostly used the quantization-based method of \cite{DNNs_Tishby2017} (on which we focus herein), they also examined replacing $\mathsf{Q}_m[T_\ell]$ with $T_\ell+\cN(\mathbf{0},\sigma^2\mathrm{I}_{d_\ell})$ (although no Gaussian noise was explicitly injected to the actual activations), as well as estimating mutual information from samples via $k$ nearest neighbor (kNN)~\cite{kraskov2004estimating} and kernel density estimation (KDE)~\cite{kolchinsky2017estimating} techniques.}

\vspace{2mm}
\subsubsection{\underline{Information plane dynamics and the effect of activation function}}

\begin{figure}[!t]
	\begin{center}
	        \includegraphics[scale = .6]{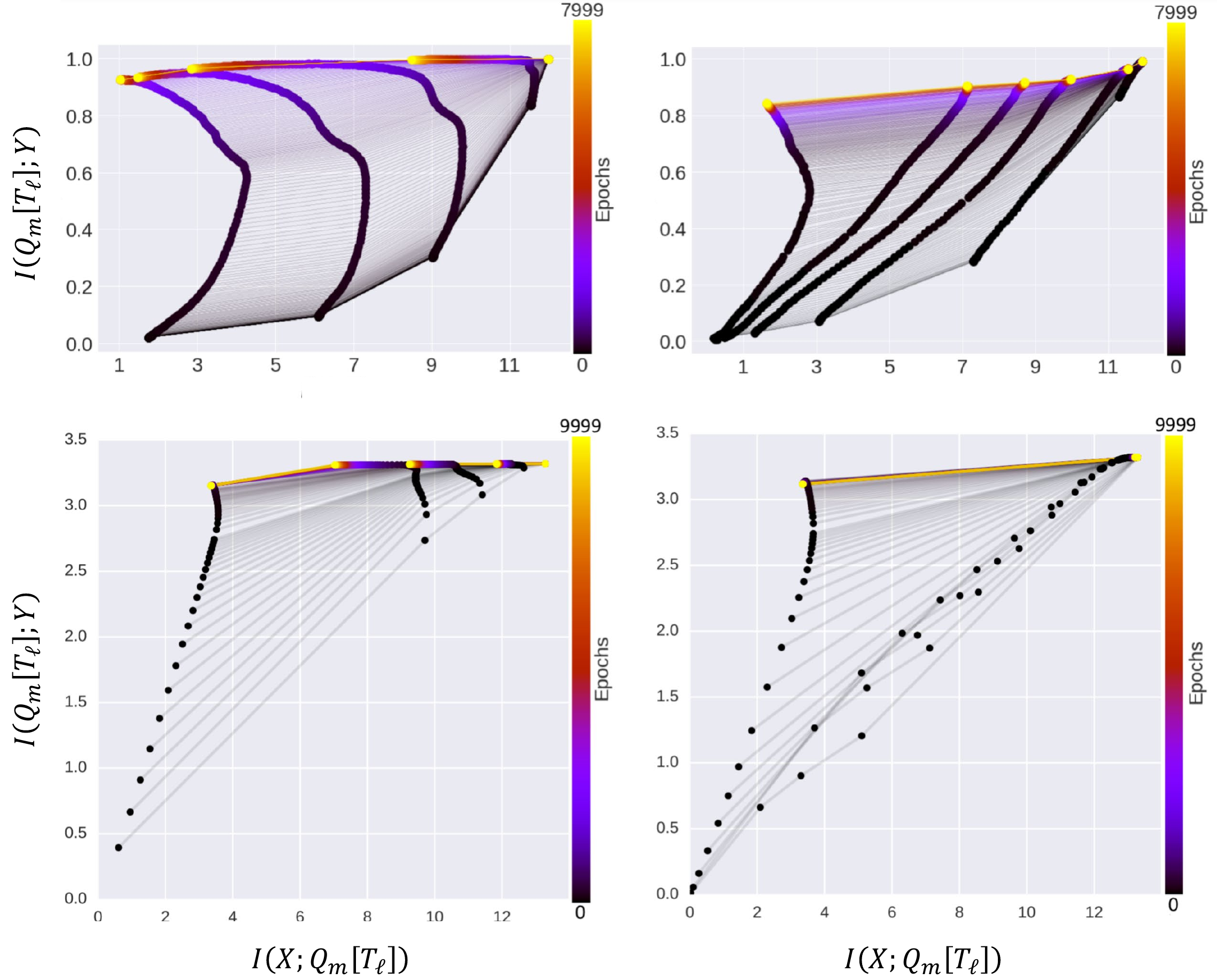}
	        \captionsetup{labelsep=none}
 	        \caption{\ [Fig. 1 from~\cite{DNNs_ICLR2018}]: IP dynamics for $\tanh$ and $\relu$ DNN classifiers (left and right, respectively) for the synthetic task from~\cite{DNNs_Tishby2017} and MNIST (top and bottom, respectively). The last layer in both $\relu$ networks (bottom row) has sigmoidal activations. While $\tanh$ networks exhibit compression of representation, no compression is observed in any of the $\relu$ layers.}\label{FIG:IP_ICLR}
	\end{center}
\end{figure}

It was argued in~\cite{DNNs_Tishby2017} that the fitting and compression phases seen in the IP are inherent to DNN classifiers trained with SGD. On the contrary,~\cite{DNNs_ICLR2018} found that the IP profile of a DNN strongly depends on the employed nonlinear activation function. Specifically, double-sided saturating nonlinearities like $\tanh$ (used in~\cite{DNNs_Tishby2017}) or $\sig$ yield a compression phase, but linear or single-sided saturating nonlinearities like $\relu(x):=\max\{0,x\}$ do not compress representations. The experiment showing this compared the IP dynamics of the same network once with $\tanh$ nonlinearities and then with $\relu$. Two tasks were examined: the synthetic experiment from~\cite{DNNs_Tishby2017} and MNIST classification. The architecture for the former was the same as in~\cite{DNNs_Tishby2017}, while the latter used a 784--1024--20--20--20--10 fully connected architecture. The last layer in both $\relu$ networks employs $\sig$ nonlinearities; all other neurons use $\relu$. Fig.~\ref{FIG:IP_ICLR} (reprinted from~\cite[Fig. 1]{DNNs_ICLR2018}) shows the IP dynamics of all four models.

First note that top-right subfigure reproduces the IP dynamics from~\cite{DNNs_Tishby2017} (compare to Fig.~\ref{FIG:phases} herein). The $\relu$ version of that network, however, does not exhibit compression, except in its last sigmoidal layer. Instead, the mutual information $I\big(X;\mathsf{Q}_m[t_\ell]\big)$ seems to monotonically increase in all $\relu$ layers. This stands in accordance with the argument of~\cite{DNNs_ICLR2018} that double-sided saturating nonlinearities can cause compression, while single-sided ones cannot. The same effect is observed for the MNIST network, by comparing its $\tanh$ version at the bottom-right with its $\relu$ version at bottom-left. Similar results were also observed for $\mathrm{soft}$-$\mathrm{sign}$ (double-sided saturation) versus $\mathrm{soft}$-$\mathrm{plus}$ (single-sided saturation) activations, given in Appendix B of~\cite{DNNs_ICLR2018}. They concluded that the choice of activation function has a significant effect on IP trajectories. In particular, the compression phase of training is caused by double-sided saturating activations, as opposed to being inherent to the learning dynamics.

\vspace{2mm}
\subsubsection{\underline{Relation between compression and generalization}} A main argument of~\cite{DNNs_Tishby2017} is that compression of representation is key for generalization. Specifically, by decreasing $I(X;T_\ell)$ (while keeping $I(T_\ell;Y)$ high), the DNN sheds information about $X$ that is irrelevant for learning $Y$, which in turn mitigates overfitting and promotes generalization.

To test this claim,~\cite{DNNs_ICLR2018} first considered deep linear networks~\cite{baldi1989neural} and leveraged recent results on generalization dynamics in the student-teacher setup~\cite{seung1992statistical,advani2017high}. In this setup, one neural network (`student') learns to approximate the output of another (`teacher'). The linear student-teacher framework with Gaussian input allows exact computation of both the generalization error and the input mutual information\footnote{To be precise, the computed quantity is $I(X;T_\ell+Z)$, for some independent Gaussian noise $Z$. The addition of $Z$ is needed, as without it $I(X;T_\ell)=\infty$ because $X$ is Gaussian and $T_\ell$ is a linear deterministic function thereof.} for a nontrivial task with interesting structure.

The exact setup is the following. Let $X\sim\cN(0,\frac{1}{d}\mathrm{I}_d)$ be an isotropic $d$-dimensional Gaussian and set the teacher network output as $Y=B_0^\top X+N_0$, where $B_0\in\mathbb{R}^d$ is the weight (column) vector and $N_0\sim \cN(0,\sigma_0^2)$ is an independent Gaussian noise. The teacher specifies a stochastic rule $P_{Y|X}$ that the student network needs to learn. Specifically, the student linear DNN is trained on a dataset generated by the teacher. Denoting the layers of the student network by $\{\mathrm{A}_\ell\}_{\ell=1}^L$, its reproduction of $Y$ is $\hat{Y}:=\mathrm{A}_L\mathrm{A}_{L-1}\cdot\mspace{-3mu}\cdot\mspace{-3mu}\cdot\mathrm{A}_1X$. Note that $X$, $Y$, $\hat{Y}$ and all the internal representation $T_\ell$ of the student network are jointly Gaussian. This allows analytic computation of generalization error and mutual information terms for any fixed parameters (i.e., at each epoch) -- see Eq. (6)-(7) of~\cite{DNNs_ICLR2018}.

Leveraging this fact, Fig. 3 of~\cite{DNNs_ICLR2018} (not shown here) compared the IP dynamics of the student linear network with a single hidden layer to the training and test errors. While the network generalized well, not compression of $I(X;T_1)$ was observed. Instead, the linear network qualitatively behaved like a $\relu$ network, presenting monotonically increasing mutual information trajectories. Building on the study of linear networks in~\cite{advani2017high}, the authors then matched the size of the student network to the number of samples, causing it to severely overfit the data. Despite now having a large generalization gap, the IP trajectory of the network did not change, still showing monotonically increasing $I(X;T_1)$ with epochs~\cite[Figs. 4A-4B]{DNNs_ICLR2018}. This produced an example of two networks with the same IP profile (no compression) but widely different generalization performance.

\begin{figure}[!t]
	\begin{center}
	        \includegraphics[scale = .5]{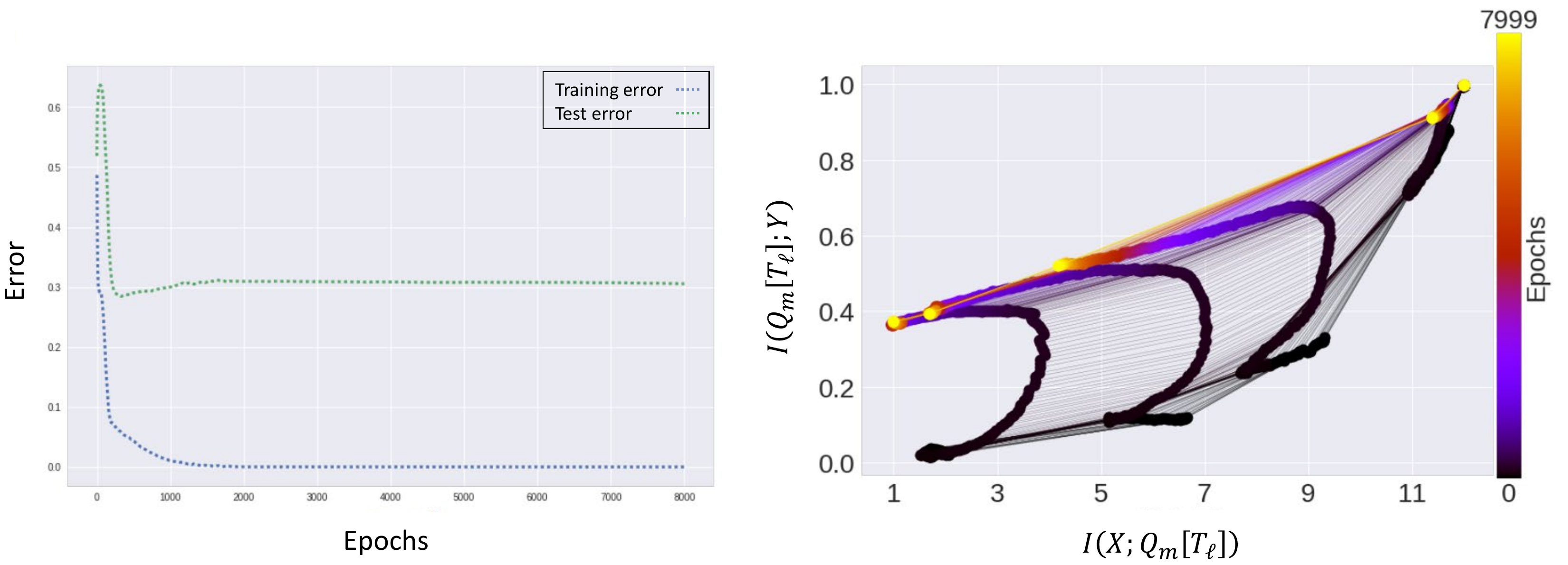}
	        \captionsetup{labelsep=none}
 	        \caption{\ [Fig. 4 from~\cite{DNNs_ICLR2018}]: Train/test errors and IP dynamics for a $\tanh$ DNN trained on $30\%$ of the synthetic dataset from ~\cite{DNNs_Tishby2017}. The network has high generalization gap, yet exhibits compression of $I\big(X;\mathsf{Q}_m[T_\ell]\big)$ in the IP.}\label{FIG:overfit}
	\end{center}
\end{figure}

Similar results were presented for nonlinear networks. The authors of~\cite{DNNs_ICLR2018} retrained the $\tanh$ network from~\cite{DNNs_Tishby2017} on the synthetic classification task therein, but using only $30\%$ of the data. This network significantly overfitted the data, resulting in a high generalization gap (left panel in Fig.~\ref{FIG:overfit}, which is reprinted from~\cite[Fig. 4]{DNNs_ICLR2018}). Nonetheless, its IP trajectories exhibit compression of $I\big(X;\mathsf{Q}_m[T_\ell]\big)$ (right panel of Fig.~\ref{FIG:overfit}). A compression phase also occurs for the original network trained on $85\%$ of the data (Figs.~\ref{FIG:phases} and~\ref{FIG:IP_ICLR}), whose test performance is much better. The main difference between the IP profiles is that the overfitted network has lower $I\big(\mathsf{Q}_m[T_\ell];Y\big)$ values, but $I\big(X;\mathsf{Q}_m[T_\ell]\big)$ compressed in both cases. 

Together, the linear and nonlinear examples dissociated compression of $I(X;T_\ell)$ (as approximated by $I\big(X;\mathsf{Q}_m[T_\ell]\big)$) and generalization: networks that compress may or may not generalize (nonlinear example), and the same applies for networks that do not compress (linear example). This suggest that the connection between compression and generalization, if exists, is not a simple causal relation. Furthermore, based on Fig.~\ref{FIG:overfit}, a direct link between $I(X;T_\ell)$ and generalization that does not involve $I(T_\ell;Y)$, seems implausible.

\vspace{2mm}
\subsubsection{\underline{Stochastic gradients drive compression}} The third claim of~\cite{DNNs_Tishby2017} revisited by~\cite{DNNs_ICLR2018} is that the randomness in SGD causes its diffusion phase, which in turn drives compression (see Section~\ref{SUBSUBSEC:SGD_dynamics}). According to this rationale, training a DNN with batch gradient decent (BGD), which updates weights using the gradient of the total error across all examples, should not induce diffusion nor result in compression. 

The authors of~\cite{DNNs_ICLR2018} trained the $\tanh$ and $\relu$ networks on the synthetic task from~\cite{DNNs_Tishby2017} with SGD and BGD. Fig. 5\footnote{We believe that Figs. 5A and 5B in~\cite{DNNs_ICLR2018} should be switched to correspond to their captions; compare Fig. 5B and Fig. 1A.} therein compares the obtained IP dynamics, showing no noticeable difference between the two training methodologies. Both SGD- and BGD-trained $\tanh$ networks present compression, while neither of the $\relu$ networks does. IP trajectories generated by SGD are shown in Fig.~\ref{FIG:IP_ICLR}; the BGD trajectories, though not presented here, looks very much alike. Interestingly,~\cite{DNNs_ICLR2018} also examined the gradient's high-to-low SNR phase transition observed in~\cite{DNNs_Tishby2017} in multiple experiments. They found that it occurs every time, regardless of the employed training method, architecture or nonlinearity type, suggesting it is a general phenomenon inherent to DNN training, though not causally related to compression of representation.

\subsection{Mutual Information Misestimation in Deterministic Networks}\label{SUBSEC:misestimation}


As discussed in Section~\ref{SUBSUBSEC:IB_vacuous}, $I(X;T_\ell)$ and $I(T_\ell;Y)$ degenerate in deterministic DNNs with strictly monotone nonlinearities. The network from~\cite{DNNs_Tishby2017} (also studied in~\cite{DNNs_ICLR2018}), whose IP dynamics are shown in Figs.~\ref{FIG:phases},~\ref{FIG:IP_ICLR} and~\ref{FIG:overfit}, is deterministic with $\tanh$ nonlinearities. Therefore, $I(X;T_\ell)=H(X)$ and $I(T_\ell;Y)=I(X;Y)$ are constant, independent of the network parameters, under the $X\sim\mathsf{Unif}(\cX_n)$ model used in these works (see Section~\ref{SUBSUBSEC:IB_vacuous}). As such, evaluating these information terms via quantization, noise injection, or estimation from samples is ill-advised. Yet, their estimates, as computed, e.g., in~\cite{DNNs_Tishby2017,DNNs_ICLR2018}, fluctuate with training epochs (that change parameter values) presenting IP dynamics. These fluctuations must therefore be a consequence of estimation errors rather than changes in mutual information. 
Indeed, quantization or noise injection are employed in~\cite{DNNs_Tishby2017,DNNs_ICLR2018} as means for performing the measurements, but they are not part of the actual network. As explained next, this creates a mismatch between the measurement model and the system being analyzed.

To simplify discussion, we focus on $I(X;T_\ell$) and its quantization-based approximation (similar reasoning applies for $I(T_\ell;Y)$ and to measurements via noise injection). Note that after quantization, the mapping from $X$ to $\mathsf{Q}_m[T_\ell]$ is no longer injective. This is since (distinct) representations $\phi_\ell(x)$ and $\phi_\ell(x')$, for $x,x'\in\cX_n$, that lie sufficiently close to one another are mapped to the same value (`quantization cell' or `bin') under $\mathsf{Q}_m$. The distances between representations are captured by the feature map $\phi_\ell(\cX_n)$, which depends on the networks parameters. As the feature map and the quantization resolution $m$ determine the distribution of $\mathsf{Q}_m[T_\ell]$, the dependence of $I\big(X;\mathsf{Q}_m[T_\ell]\big)$ on the network parameters becomes clear. This results in a parameter-dependent estimate $I\big(X;\mathsf{Q}_m[T_\ell]\big)$ of the parameter-independent quantity $I(X;T_\ell)$, which is undesirable. Strictly speaking, there is nothing here to estimate: since $X\sim\mathsf{Unif}(\cX_n)$ (as assumed in~\cite{DNNs_Tishby2017,DNNs_ICLR2018}) and $\phi_\ell$ is injective from $\cX_n$ to $\phi_\ell(\cX_n)$, the true $I(X;T_\ell)$ value is~$\log n$.

\begin{figure}[!t]
	\begin{center}
    \includegraphics[width=\textwidth]{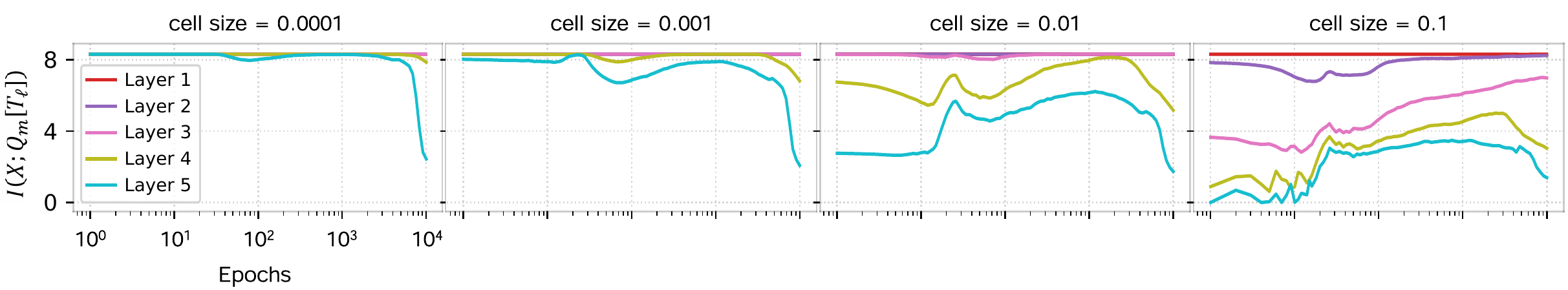}
    \captionsetup{labelsep=none}
    \caption{\ [Fig. 1 from~\cite{ICML_Info_flow2019}]: $I\big(X;\mathsf{Q}_m[T_\ell]\big)$ vs. epochs for different quantization cell (bin) sizes and the model in~\cite{DNNs_Tishby2017}, where $X$ is uniformly distributed over a $2^{12}$-sized empirical dataset. The curves converge to $H(X)=\ln(2^{12})\approx 8.3$ for small bins.}
    \label{Fig:Binning_tanh}
    \end{center}
\end{figure}

Recalling \eqref{EQ:MI_quantization}, we expect that $I\big(X;\mathsf{Q}_m[T_\ell]\big)\to H(X)=\log n$ as $m\to 0$. For nonnegligible $m>0$, the value of $I\big(X;\mathsf{Q}_m[T_\ell]\big)$ strongly depends on the quantization parameter. 
However, since no quantization is present in the actual network (see \eqref{EQ:DNN_def}), the value of $m$ is arbitrary and chosen by the user. Thus, the measured $I\big(X;\mathsf{Q}_m[T_\ell]\big)$ reveal more about the estimator and the dataset than about the true mutual information, which is a global property of the underlying joint distribution. 
The dependence of $I\big(X;\mathsf{Q}_m[T_\ell]\big)$  on $m$ is illustrated in Fig.~\ref{Fig:Binning_tanh} (reprinted from~\cite[Fig. 1]{ICML_Info_flow2019}), showing that widely different profiles can be obtained by changing $m$. The leftmost subfigure also shows how the quantized mutual information approaches $H(X)=\log n$ as $m$ shrinks. Recalling that the dataset size in~\cite{DNNs_Tishby2017} is $n=2^{12}$ and taking logarithms to the base of $e$, we see that for $m=10^{-4}$, $I\big(X;\mathsf{Q}_m[T_\ell]\big)\approx \log2^{12}\approx 8.3$, for all hidden layers $\ell=1,\ldots,5$ and across (almost) all epochs of training.

In summary, while $I\big(X;\mathsf{Q}_m[T_\ell]\big)$ fails to capture the true (vacuous) mutual information value, it encodes nontrivial information about the feature maps. The compression phase observed in~\cite{DNNs_Tishby2017} is in fact a property of $I\big(X;\mathsf{Q}_m[T_\ell]\big)$, rather than $I(X;T_\ell)$, driven by the evolution of internal representations. Notably, for any of the quantization resolutions shown in Fig.~\ref{Fig:Binning_tanh}, at least for the last hidden layer, $I\big(X;\mathsf{Q}_m[T_5]\big)$ always undergoes a decrease towards the end of training. This raises the questions of what is the underlying phenomenon inside the internal representation spaces that causes this behavior. As was shown in~\cite{ICML_Info_flow2019}, the answer turns out to be clustering. The next section focuses on the methodology and the results of~\cite{ICML_Info_flow2019} that led to this conclusion.


\subsection{Noisy Deep Networks and Relation to Clustering of Representations}\label{SUBSEC:IB_clustering}

While $I(X;T_\ell)$ is vacuous in deterministic DNNs, the compression phase that its estimate  $I\big(X;\mathsf{Q}_m[T_\ell]\big)$ undergoes during training seems meaningful. To study this phenomenon,~\cite{ICML_Info_flow2019} developed a rigorous framework for tracking the flow of information in DNNs. Specifically, they proposed an auxiliary `noisy' DNN setup, under which the map $X\mapsto T_\ell$ is a \emph{stochastic parameterized channel}, whose parameters are the network's weights and biases. This makes $I(X;T_\ell)$ over such networks, a meaningful system-dependent~quantity. The authors of~\cite{ICML_Info_flow2019} then proposed a provably accurate estimator of $I(X;T_\ell)$ and studied its evolution during training. Although not covered in detail herein, additional ways to make the IB non-vacuous (beyond noising the activations) include: adding noise to the weights~\cite{Achille2018,achille2019information}, changing the objective~\cite{strouse2017deterministic}, and changing the information measure~\cite{cvitkovic2019minimal,wickstrom2019information}.


\vspace{2mm}

\subsubsection{\underline{Noisy DNNs}} The definition of a noisy DNN replaces $T_\ell=\phi_\ell(X)$ (see \eqref{EQ:DNN_def}) with $T_\ell^{(\sigma)}:=T_\ell+Z_\ell^{(\sigma)}$, where $\big\{Z_\ell^{(\sigma)}\big\}_{\ell=1}^L$ are independent isotropic $d_\ell$-dimensional Gaussian vectors of parameter $\sigma>0$, i.e., $Z_\ell^{(\sigma)}\sim\Gauss:=\mathcal{N}(0,\sigma^2\mathrm{I}_d)$. In other words, i.i.d. Gaussian noise is injected to the output of each hidden neuron. The noise here is intrinsic to the system, i.e, the network is trained with the noisy activation values. This stands in contrast to~\cite{DNNs_Tishby2017} and~\cite{DNNs_ICLR2018}, where binning or noise injection were merely a part of the measurement (of mutual information) model. Intrinsic noise as in~\cite{ICML_Info_flow2019} ensures that the characteristics of \emph{true} information measures over the network are tied to the network's dynamics and the representations it is learning. Furthermore, the isotropic noise model relates $I\big(X;T_\ell^{(\sigma)}\big)$ to $I\big(X;\mathsf{Q}_m[T_\ell]\big)$ when $\sigma$ is of the order of the quantization cell side length. This is important since it is the compression of the latter that was observed in preceding works. 

To accredit the noisy DNN framework,~\cite{ICML_Info_flow2019} empirically showed that it forms a reasonable proxy of deterministic networks used in practice. Namely, when $\sigma>0$ is relatively small (e.g., of the order of $10^{-2}$ for a $\tanh$ network), it was demonstrated that noisy DNNs not only perform similarly to deterministic ones, but also that the representations learned by both systems are closely related.


\vspace{2mm}

\subsubsection{\underline{Mutual information estimation}} 

Adopting $\big(I\big(X;T_\ell^{(\sigma)}\big),I\big(T_\ell^{(\sigma)};Y\big)\big)$ as the figure of merit in noisy DNNs, one still faces the task of evaluating these mutual information terms. Mutual information is a functional of the joint distribution of the involved random variables. While, $T_\ell=\phi_\ell(X)+Z_\ell^{(\sigma)}$ and $\phi_\ell$ is specified by the (known) DNN model, the data-label distribution $P_{X,Y}$ is unknown and we are given only the dataset $\cD_n$. Statistical learning theory treats the elements of $\cD_n$ as i.i.d. samples from $P_{X,Y}$. Under this paradigm, evaluating the mutual information pair of interest enters the realm of statistical estimation~\cite{paninski2004estimating,kraskov2004estimating,valiant2013estimating,han2015adaptive,han2017optimal,Hero_EDGE2018,goldfeld2019convergence,belghazi2018mutual,chung2019neural}. However, mutual information estimation from high-dimensional data is a notoriously difficult~\cite{Paninski2003}. Corresponding error convergence rates (with $n$) in high-dimensional settings are too slow to be useful in practice.


Nevertheless, by exploiting the known distribution of the injected noise,  \cite{ICML_Info_flow2019} proposed a rate-optimal estimator of $I\big(X;T_\ell^{(\sigma)}\big)$ that scales significantly better with dimension than generic methods (such as those mentioned above). This was done by developing a forward-pass sampling technique that reduced the estimation of $I\big(X;T_\ell^{(\sigma)}\big)$ to estimating differential entropy under Gaussian noise as studied in \cite{goldfeld2019convergence} (see also~\cite{Goldfeld_estimation_ICSEE2019,Goldfeld_estimation_ISIT2019,goldfeld2020limit}). Specifically, the latter considered  estimating the differential entropy $h(S+Z)=h(P\ast\Gauss)$ based on `clean' samples of $S\sim P$, where $P$ belongs to a nonparametric distribution class, and knowledge of the distribution of $Z\sim\Gauss$, which is independent of $S$. Here $(P\ast Q)(\mathcal{A}):=\int\int\mathds{1}_{\mathcal{A}}(x+y)\dd P(x)\dd Q(y)$ is the convolution of two probability measures $P$ and $Q$, and $\mathds{1}_\mathcal{A}$ is the indicator function of $\mathcal{A}$. 

The reduction of mutual information estimation to estimating $h(S+Z)$ uses the decomposition
\begin{equation}
    I\big(X;T_\ell^{(\sigma)}\big)=h(T_\ell^{(\sigma)}\big)-\int h\big(T_\ell^{(\sigma)}\big|X=x\big)\dd P_X(x),
\end{equation}
along with the fact that $T_\ell^{(\sigma)}=T_\ell+Z_\ell^{(\sigma)}$, where $T_\ell=\sigma(\mathrm{A}_\ell\phi_{\ell-1}(X)+b_\ell)$, is easily sampled via the forward-pass of the network (see~\cite[Section IV]{goldfeld2019convergence} and~\cite[Section III]{ICML_Info_flow2019} for more details).\footnote{Samples for estimating the conditional differential entropy terms $h\big(T_\ell^{(\sigma)}\big|X=x\big)$, for $x\in\cX$, are obtained by feeding the network with $x$ multiple times and reading $T_\ell^{(\sigma)}$ values corresponding to different noise realizations.} Building on~\cite{goldfeld2019convergence}, the employed estimator for $h(P\ast\Gauss)$ was $\hat{h}(S^n,\sigma):=h(\hat{P}_{S^n}\ast\Gauss)$, where $\hat{P}_{S^n}:=\frac{1}{n}\sum_{i=1}^n\delta_{Si}$ is the empirical distribution of the i.i.d. sample $S^n:=(S_1,\ldots,S_n)$ of $S\sim P$. The estimation risk of $\hat{P}_{S^n}$ over the class of all compactly supported~\cite[Theorem 2]{goldfeld2019convergence} or sub-Gaussian~\cite[Theorem 3]{goldfeld2019convergence} $d$-dimensional distributions $P$ scales as $c^d n^{-\frac{1}{2}}$, for an explicitly characterized numerical constant $c$. Matching impossibility results showed that this rate is optimal both in $n$ and $d$. By composing multiple differential entropy estimators (for $h\big(T_\ell^{(\sigma)}\big)$ and each $h\big(T_\ell^{(\sigma)}\big|X=x\big)$, $x\in\cD_n$), an estimator $\hat{I}(\cD_n,\sigma)$ of $I\big(X;T_\ell^{(\sigma)}\big)$ was constructed. Its absolute-error estimation risk over, e.g., DNNs with $\tanh$ nonlinearities scales as follows.

\begin{proposition}[Mutual Information Estimator~\cite{goldfeld2019convergence}]\label{PROP:MI_True_Data_Dist}
For the described estimation setup, we have
\begin{align*}
    \sup_{P_X}\mathbb{E} &\left|I(X;T)-\hat{I}_{\mathsf{Input}}\p{X^n,\hat{h},\sigma}\right|\leq \frac{8c^{d_\ell}+d_\ell\log\left(1+\frac{1}{\sigma^2}\right)}{4\sqrt{n}},
\end{align*}
where $c$ is a constant independent of $n$, $d$ and $P_X$, which is explicitly characterized in~\cite[Equation (61)]{goldfeld2019convergence}.
\end{proposition}
Notably, the right-hand side depends exponentially on dimensions, which limits the dimensionality of experiments for which the bound in non-vacuous. This limitation is inherent to the considered estimation problem, as~\cite{goldfeld2019convergence} proved that the sample complexity of \emph{any} good estimator depends exponentially on $d_\ell$.


\vspace{2mm}
\subsubsection{\underline{Empirical study and relation to clustering}}

\begin{figure}[!t]
    \centering
    \setlength{\tabcolsep}{0pt}
    \includegraphics[width=\textwidth]{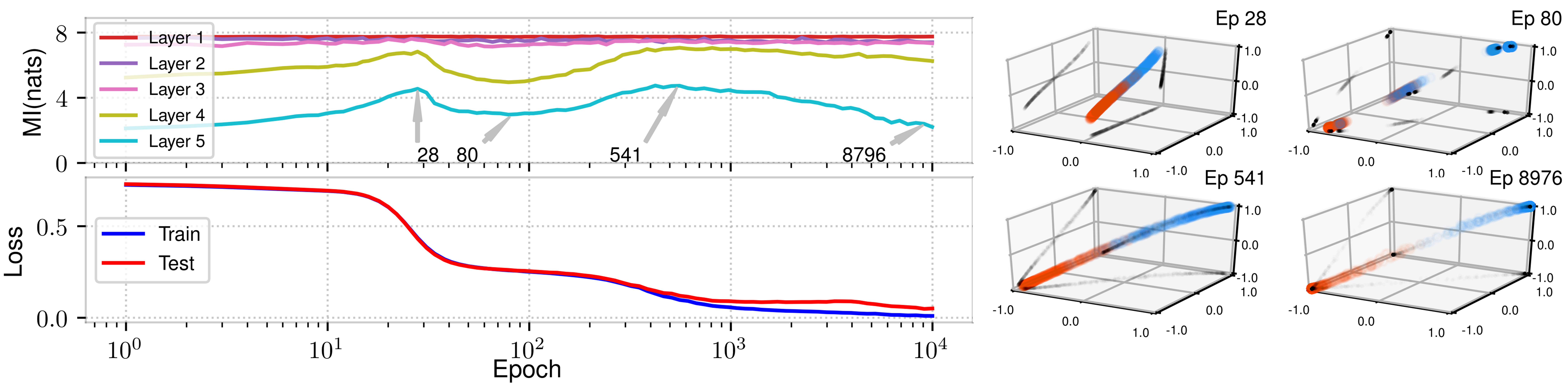}
\caption{$I\big(X;T_\ell^{(\sigma)}\big)$ evolution and training/test losses across training for a noisy version of the DNN from~\cite{DNNs_Tishby2017}. Scatter plots show the input samples as represented by the $5^{\mbox{\scriptsize{th}}}$ hidden layer (colors designate classes) at the arrow-marked epochs.}\label{FIG:clustering}
\end{figure}

The developed toolkit enabled the authors of~\cite{ICML_Info_flow2019} to accurately track $I\big(X;T_\ell^{(\sigma)}\big)$ during training of (relatively small) noisy DNN classifiers. For the synthetic experiment from~\cite{DNNs_Tishby2017} with $\tanh$ activations,~\cite{ICML_Info_flow2019} empirically showed that $I\big(X;T_\ell^{(\sigma)}\big)$ indeed undergoes a long-term compression phase (see Fig.~\ref{FIG:clustering}, which is reprinted from~\cite[Fig. 5(a)]{ICML_Info_flow2019}). To reveal the geometric mechanism driving this phenomenon, they related $I\big(X;T_\ell^{(\sigma)}\big)$ to data transmission over AWGN channels. The mutual information $I\big(X;T_\ell^{(\sigma)}\big)$ can be viewed as the aggregate number of bits (reliably) transmittable over an AWGN channel using an input drawn from the latent representation space. As training progresses, the hidden representations of equilabeled inputs cluster together, becoming increasingly indistinguishable at the channel's output, thereby decreasing $I\big(X;T_\ell^{(\sigma)}\big)$. To test this empirically,~\cite{ICML_Info_flow2019} contrasted $I\big(X;T_\ell^{(\sigma)}\big)$ trajectories with scatter plots of the feature map $\phi_\ell(\cX_n)$. Remarkably, compression of mutual information and clustering of latent representations clearly corresponded to one another. This can be seen in Fig.~\ref{FIG:clustering} by comparing the scatter plots on the right to the arrow-marked epochs in the information flow trajectory on the left. 

Next,~\cite{ICML_Info_flow2019} accounted for the observations from~\cite{DNNs_Tishby2017,DNNs_ICLR2018} of compression in deterministic DNNs. It was demonstrated that while the quantization-based estimator employed therein fails to capture the true (constant/infinite) mutual information, it does serve as a measure of clustering. Indeed, for a deterministic network we have
$$I\big(X;\mathsf{Q}_m[T_\ell]\big)=H\big(\mathsf{Q}_m[T_\ell]\big),$$
where $\mathsf{Q}_m$ partitions the dynamic range (e.g., $[-1,1]^{d_\ell}$ for a $\tanh$ layer) into $m^{d_\ell}$ cells. When hidden representations are spread out, many cells are non-empty, each having some small positive probability mass. Conversely, for clustered representations, the distribution is concentrated on a small number of cells, each with relatively high probability. Recalling that the uniform distribution maximizes Shannon entropy, we see that reduction in $H\big(\mathsf{Q}_m[T_\ell]\big)$ corresponds to tighter clusters in the latent representation space.

The results of~\cite{ICML_Info_flow2019} identified the geometric clustering of representations as the fundamental phenomenon of interest, while elucidating some of the machinery DNNs employ for learning. Leveraging the clustering perspective,~\cite{ICML_Info_flow2019} also provided evidence that compression and generalization may \emph{not} be causally related. Specifically, they constructed DNN examples that generalize better when tight clustering is actively suppressed during training (compare Figs. 5(a) and 5(b) from~\cite{ICML_Info_flow2019}). This again showed that the relation between compression of mutual information and generalization is not a simple one, warranting further study. More generally, there seems to be a mismatch between the IB paradigm and common DL practice that mainly employs deterministic NNs, under which information measures of interest tend to degenerate. It remains unclear how to bridge this gap. 





\section{Summary and Concluding Remarks}\label{SEC:sumamry}

This tutorial surveyed the IB problem, from its information-theoretic origins to the recent impact it had on ML research. After setting up the IB problem, we presented its relations to MSSs, discussed operational interpretations, and covered the Gaussian IB setup. Together, these components provide background and context for the recent framing of IB as an objective/model for DNN-based classification. After describing successful applications of the IB framework as an objective for learning classifiers~\cite{alemi2017deep,achille2018information} and generative models \cite{higgins2017beta}, we focused on the IB theory for DL~\cite{tishby_DNN2015,DNNs_Tishby2017} and the active research area it inspired. The theory is rooted in the idea that DNN classifiers inherently aim to learn representations that are optimal in the IB sense. This novel perspective was combined in~\cite{DNNs_Tishby2017} with an empirical case-study to make claims about phase transitions in optimization dynamics, computational benefits of deep architectures, and relations between generalization and compressed representations. Backed by some striking empirical observations (though only for a synthetic classification task), the narrative from~\cite{DNNs_Tishby2017} ignited a series of followup works aiming to test its generality. 

We focused here on works that contributed to different aspect of modern IB research. Our starting point is~\cite{DNNs_ICLR2018}, that revisited the observations from~\cite{DNNs_Tishby2017} with a thorough empirical analysis. The experiments from~\cite{DNNs_ICLR2018} were designed to test central aspects of the IB theory for DL; their final conclusion was that the empirical findings from~\cite{DNNs_Tishby2017} do not hold in general. We then examined theoretical facets applying the (inherently stochastic) framework of IB to deterministic DL systems. Covering contributions from~\cite{amjad2019learning} and~\cite{ICML_Info_flow2019}, caveats in measuring information flows in deterministic DNNs were explained. In a nutshell, key information measures degenerate over deterministic networks, becoming either constant (independent of the network's parameters) or infinite, depending on the modeling of the input feature. Either way, such quantities are vacuous over deterministic networks. 

We then described the remedy proposed in~\cite{ICML_Info_flow2019} to the `vacuous information measures' issue. Specifically, that work presented an auxiliary stochastic DNN framework over which the considered mutual information terms are meaningful and parameter-dependent. Using this auxiliary model, they demonstrated that compression of $I(X;T)$ over the course of training, is driven by clustering of equilabeled samples in the representation space of $T$. It was then shown that a similar clustering process occurs during training of deterministic DNN classifiers. Circling back to the original observations of compression~\cite{DNNs_Tishby2017} (see also~\cite{DNNs_ICLR2018}), the authors of \cite{ICML_Info_flow2019} demonstrated that the measurement techniques employed therein in fact track clustering of samples. This clarified the geometric phenomena underlying the compression of mutual information during training. Still, many aspects of the IB theory for DL remain puzzling, awaiting further exploration.

\newpage

\bibliographystyle{IEEEtran}
\bibliography{ref}
	
\end{document}